\documentclass{article}

\usepackage[preprint]{neurips_2026}

\usepackage[utf8]{inputenc}
\usepackage[T1]{fontenc}

\usepackage{amsmath,amsfonts,bm}

\def\eqref#1{equation~\ref{#1}}

\def\1{\bm{1}}

\DeclareMathAlphabet{\mathsfit}{\encodingdefault}{\sfdefault}{m}{sl}
\SetMathAlphabet{\mathsfit}{bold}{\encodingdefault}{\sfdefault}{bx}{n}

\usepackage[colorlinks,linkcolor=black,citecolor=blue!70!black,urlcolor=blue!70!black]{hyperref}
\usepackage{url}
\usepackage{graphicx}

\usepackage{multirow}
\usepackage{booktabs}
\usepackage{colortbl}
\usepackage{wrapfig}
\usepackage{caption}
\usepackage{threeparttable}
\usepackage{enumitem}
\usepackage{nicefrac}

\usepackage[capitalize,noabbrev]{cleveref}

\usepackage{xcolor}

\usepackage{amsmath}
\usepackage{mathtools}

\usepackage{tcolorbox}
\tcbuselibrary{breakable}

\usepackage{placeins}
\usepackage{float}

\usepackage{algorithm}
\usepackage{algorithmic}

\usepackage{xspace}

\newcommand{{\method}}{OpenSIR\xspace}

\definecolor{lightblue}{RGB}{173, 216, 230}
\definecolor{mypurple1}{RGB}{143, 94, 255}
\definecolor{myblue1}{RGB}{59, 76, 206}
\definecolor{myred1}{RGB}{190, 90, 75}
\definecolor{myyellow1}{RGB}{255, 192, 0}
\definecolor{mygreen1}{RGB}{160, 194, 128}
\definecolor{nodiversity}{RGB}{255, 107, 107}  
\definecolor{withdiversity}{RGB}{78, 205, 196}  
\definecolor{mathcolor}{RGB}{255, 215, 0}       
\definecolor{gsmcolor}{RGB}{128, 0, 128}        

\definecolor{impcolor}{RGB}{108, 144, 232}
\definecolor{deccolor}{RGB}{128, 128, 128}

\newcommand{\cimp}[2]{%
  #1\rlap{\textsuperscript{\color{impcolor}#2}}%
}

\newcommand{\bcimp}[2]{%
  \textbf{#1\rlap{\textsuperscript{\color{impcolor}#2}}}%
}

\title{OpenSIR: Open-Ended Self-Improving Reasoner}

\author{%
  \begin{tabular}{@{}c@{\hspace{2.5em}}c@{\hspace{2.5em}}c@{}}
    Wai-Chung Kwan$^{1}$ &
    Joshua Ong Jun Leang$^{1,2}$ &
    Pavlos Vougiouklis$^{3}$ \\
    Jeff Z. Pan$^{1}$ &
    Marco Valentino$^{4}$ &
    Pasquale Minervini$^{1,5}$
  \end{tabular} \\
  \normalfont $^{1}$University of Edinburgh \qquad
  $^{2}$Imperial College London \\
  \normalfont $^{3}$Huawei Technologies Research \& Development (UK) Limited \\
  \normalfont $^{4}$University of Sheffield \qquad
  $^{5}$Miniml.AI \\
  \normalfont \texttt{\{wkwan, p.minervini\}@ed.ac.uk}
}

\begin{document}

\maketitle

\vspace{-1em}
\makeatletter\if@preprint\makeatother
  \begin{center}
    {\fontsize{11pt}{13pt}\selectfont
      \href{https://github.com/EdinburghNLP/OpenSIR}{\raisebox{-0.1em}{\includegraphics[height=1.2em]{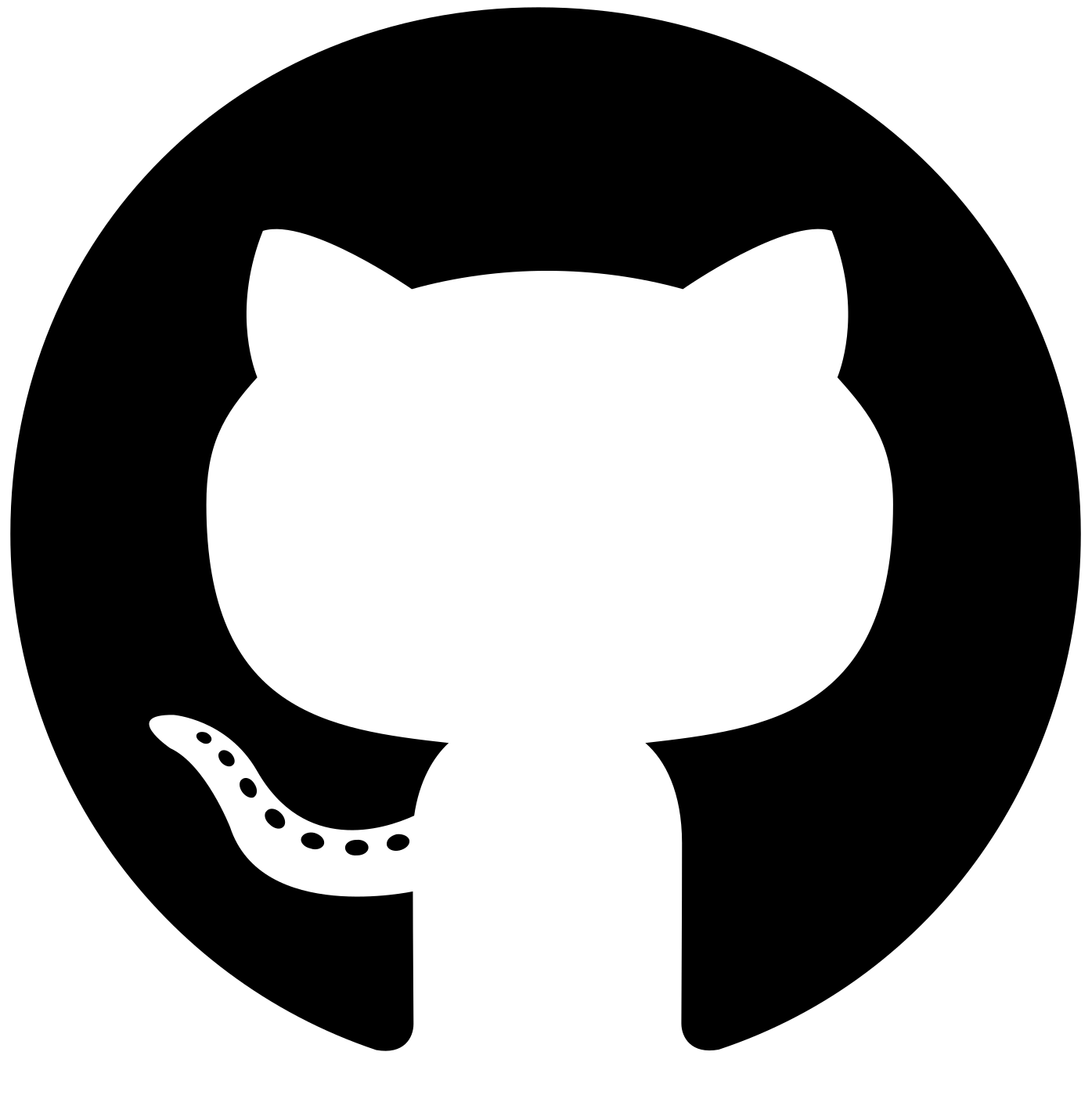}}~Code}}
  \end{center}
  \makeatletter\fi\makeatother
\vspace{1em}

\begin{abstract}
  Recent advances in large language model (LLM) reasoning through reinforcement learning rely on annotated datasets for verifiable rewards, which may limit models' ability to surpass human-level performance.
  While self-play offers a promising alternative, prior methods yield only marginal or even negative gains on post-trained models because they generate problems that cluster around familiar concepts rather than discovering novel ones.
  We introduce \textbf{Open}-Ended \textbf{S}elf-\textbf{I}mproving \textbf{R}easoner ({\method}), a self-play framework in which a single LLM alternates teacher and student roles to generate and solve novel problems without external verifiers or annotated data.
  Starting from a single seed problem, {\method} sustains open-ended exploration through diversity rewards that push the model toward unfamiliar concepts and difficulty calibration that keeps problems learnable.
  Across seven math benchmarks, {\method} consistently improves all models, averaging +3.6 points on instruction models and +3.1 on reasoning models, while recent self-play baselines yield marginal or even negative gains; starting from a single trivial seed, it also surpasses GRPO baselines trained on over 7K annotated examples.
  Despite training only on self-generated math, {\method} is the only self-play method that transfers to general reasoning, improving by at least +4.4 points on reasoning models.
\end{abstract}

\section{Introduction}
Reinforcement learning with verifiable rewards (RLVR) drives recent advances in LLM reasoning.
Recent works on DeepSeek-R1 \citep{deepseek-ai_deepseek-r1_2025} and OpenAI o1~\citep{openai_learning_2024} have shown that large-scale reinforcement learning improves reasoning capabilities. Yet, these methods require extensive human-annotated data for reward signals, which bottleneck scalability and potentially limit performance to human-level~\citep{DBLP:conf/icml/0001DPBMSSR24}.
One promising direction to address these fundamental limitations is to generate synthetic training data through self-play, which demonstrated remarkable success in various games~\citep{silver_mastering_2016, silver_mastering_2017-1,brown_superhuman_2019,meta_fundamental_ai_research_diplomacy_team_fair_human-level_2022}, allowing systems to exceed human-level performance by learning from unambiguous reward signals~\citep{silver_mastering_2017-1,meta_fundamental_ai_research_diplomacy_team_fair_human-level_2022}.

Applying self-play to mathematical reasoning reveals a fundamental obstacle: unlike games with clear rules and winners, generated mathematics problems lack ground-truth answers to provide feedback signals.
Existing approaches work around this by importing external verifiers, such as compilers~\citep{pourcel_aces_2024,zhao_absolute_2025} or game rules~\citep{liu_spiral_2025}, or by relying on heuristic agreement signals such as majority voting with repetition penalties, as in R-Zero~\citep{huang_r-zero_2025}.
Absolute Zero~\citep{zhao_absolute_2025} and R-Zero~\citep{huang_r-zero_2025} take this approach furthest, as both generate and solve problems within a single framework without relying on external verifiers.
Yet neither was evaluated on post-trained models, the very setting where self-play is most needed to push beyond what human-curated data has already provided.
When applied to post-trained models, their gains are typically marginal and can even turn negative (\S\ref{sec:main-results}).
This failure points to a deeper limitation: neither method supports open-ended learning~\citep{bauer_human-timescale_2023,hughes_open-endedness_2024}---the capacity to continuously generate and pursue genuinely novel challenges without external supervision.
Without it, the teacher collapses toward familiar problem templates and ceases to produce new learning signal, rendering self-play ineffective for models that have already mastered human-curated data.
\begin{figure}[t!]
  \centering
  \includegraphics[width=.8\textwidth]{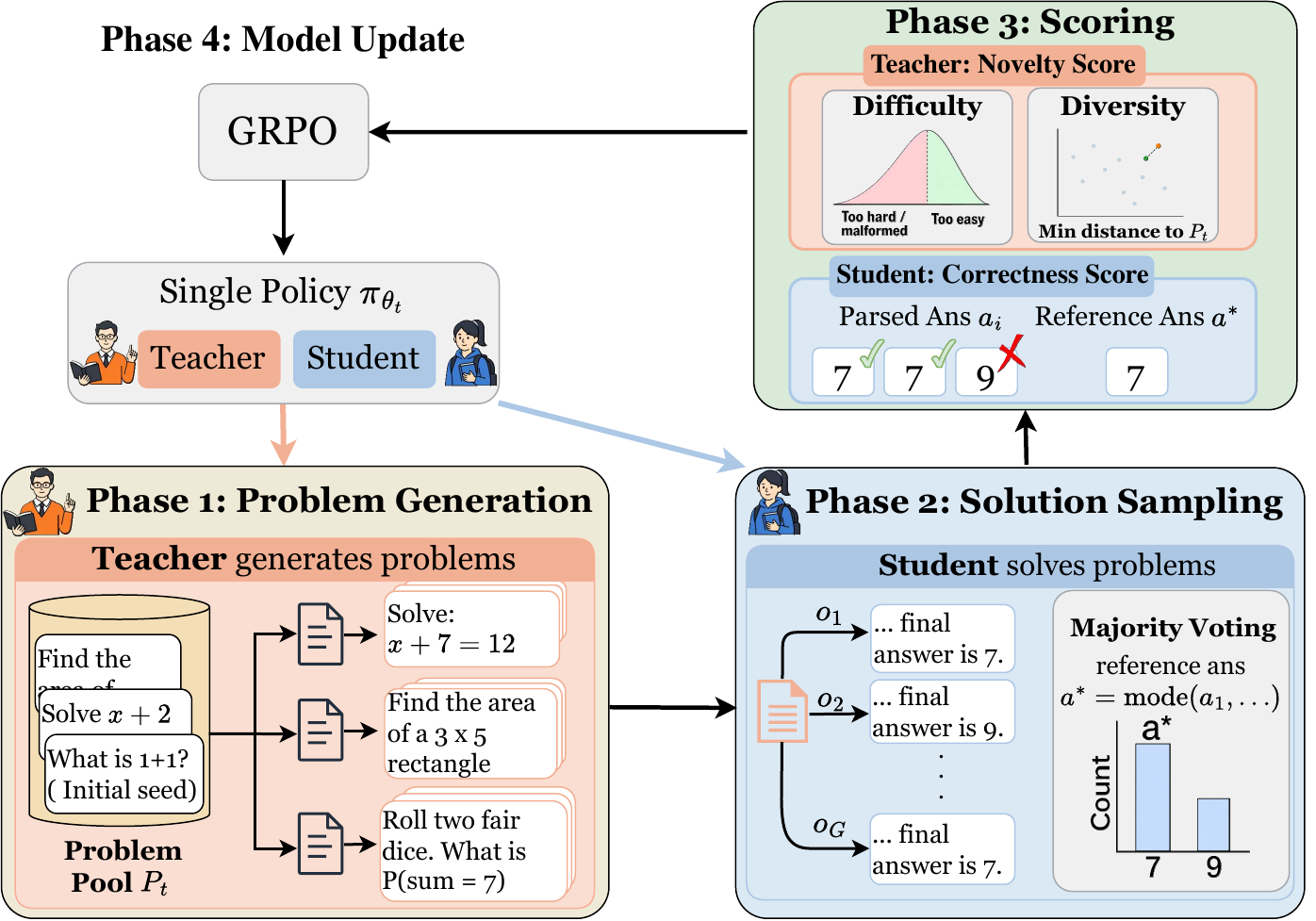}
  \caption{Overview of the {\method} framework.
    A single policy $\pi_\theta$ alternates between generating and solving novel problems without external supervision.
    Each training iteration consists of \textbf{problem generation}, \textbf{solution sampling}, \textbf{scoring}, and \textbf{model update}.
    Novelty is captured through both \emph{difficulty} and \emph{diversity}: problems must be challenging yet solvable, and they must explore new concepts.
    These dimensions together drive open-ended self-improvement in the LLM reasoning ability.
  } \label{fig:main}
\end{figure}
To address this limitation, we present \textbf{Open}-ended \textbf{S}elf-\textbf{I}mproving \textbf{R}easoner ({\method}), a self-play framework in which a single policy $\pi_\theta$ alternates between teacher and student roles to continuously generate and solve novel problems without external supervision. {\method} takes a concrete step toward open-ended learning~\citep{bauer_human-timescale_2023,hughes_open-endedness_2024} by sustaining novelty-driven self-improvement from a single trivial seed and building broadly transferable reasoning skills (\S\ref{sec:main-results}).
  {\method} captures novelty through two reward components.
A \emph{difficulty} reward, computed from consistency and solution length across multiple solution attempts, selects problems that are appropriately challenging, while an embedding-based \emph{diversity} reward promotes problems that explore semantically new mathematical concepts.
Together, these rewards continually expand the model's capacity, sustaining the novel learning signal that prior self-play methods fail to maintain.

Across seven math and three general reasoning benchmarks, existing self-play methods yield negligible gains on instruction models (up to $+$0.87) and actively degrade reasoning models (up to $-$1.93). {\method} consistently improves all models, averaging $+$3.35 on maths and up to $+$4.79 on general reasoning, highlighting the need for novelty to sustain learning in well post-trained models.
Starting from a single trivial seed problem, these results surpass even GRPO baselines trained on over 7,000 human-annotated examples.
Our analysis shows that the diversity reward is central to this behaviour: it nearly doubles concept coverage and helps the gains transfer beyond math to general reasoning.
We also find that teacher and student must co-evolve to keep problem difficulty calibrated, while the solvability reward prevents the model from chasing harder but increasingly invalid problems that hurt learning.

Our contributions are as follows:
\begin{enumerate}
  \item We show that existing self-play methods fail on well post-trained models and that problem diversity is critical for sustained improvement.
  \item We propose \method, the first self-play framework to enable open-ended self-improvement by jointly optimising problem difficulty and diversity, bootstrapping from a single trivial seed without external supervision.
  \item We demonstrate that {\method} succeeds where prior self-play methods fail, reliably improving well post-trained models with gains that transfer to general reasoning and outperform GRPO baselines trained on over 7,000 annotated examples.
\end{enumerate}

\section{Open-Ended Self-Improving Reasoner}
Figure \ref{fig:main} illustrates the Open-Ended Self-Improving Reasoner ({\method}), a self-play framework in which a policy $\pi_\theta$ learns to both generate and solve novel mathematical problems without external supervision.
We use reinforcement learning to optimise two roles within one policy: the \emph{teacher}, which creates new problems, and the \emph{student}, which solves them.
This open-ended approach enables the policy to bootstrap its learning and discover new and diverse challenges without annotated data.
Each training iteration involves four phases (Algorithm~\ref{alg:opensir}):
\begin{enumerate}[leftmargin=*]
  \item \textbf{Problem generation} (\S\ref{sec:problem-generation}): The teacher proposes new problems by conditioning on reference problems from an accumulated pool of previously generated problems;
  \item \textbf{Solution sampling} (\S\ref{sec:solution-sampling}): The student attempts multiple solutions per problem, with majority voting determining the reference answer and solve rate measuring reliability;
  \item \textbf{Scoring} (\S\ref{sec:scoring}): We compute novelty scores for the teacher's generated problems and correctness scores for the student's solutions; and
  \item \textbf{Model update} (\S\ref{sec:problem-selection}): We update the policy's parameters with role-specific rewards using the problem-solution pairs selected by the novelty scores.
\end{enumerate}
In {\method}, we define novelty along two dimensions that together drive continuous open-ended learning.
First, problems must have an appropriate level of difficulty.
It should be challenging enough to promote learning but solvable enough to provide reliable training signals.
Second, problems must explore diverse concepts, preventing the model from repeating learning on familiar concepts.
This two-dimensional view of novelty ensures the model continuously expands both the depth and breadth of its mathematical reasoning abilities.
\subsection{Problem Generation}
\label{sec:problem-generation}
At each iteration $t$, the policy $\pi_\theta$ generates $k$ groups of $G$ problems each, denoted as $q_{1:G}$ within each group, for a total of $M = k \times G$ problems.
To generate these problems, we sample $k$ reference problems from a pool $\mathcal{P}_{t-1}$ of accumulated problems from previous iterations, where each reference problem serves as a seed for generating $G$ new problems.
Each generated problem must explicitly include the mathematical concepts required for its solution.
Problems with invalid formats are filtered out, and valid problems proceed to the solution-sampling phase.
We initialise the problem pool $\mathcal{P}_{0}$ with a single trivial problem (``What is 1+1?'').

\subsection{Solution Sampling}
\label{sec:solution-sampling}
Let $a_j$ denote the parsed answer from solution attempt $o_j$.
We select the most common answer across attempts as the reference answer $a^*$.
We then compute the \emph{solve rate} for each problem to determine the reliability of the answers.
For brevity, we denote $s_{q_i} = \text{SolveRate}(q_i)$ when referring to the solve rate of problem $q_i$.
\begin{equation} \label{eq:solverate}
  \text{SolveRate}(q_i) = \frac{\text{count}(a^*)}{G}, \quad a^* = \underset{a \in a_{1:G}}{\arg\max} \; \text{count}(a)
\end{equation}
In \cref{eq:solverate}, $\text{count}(a)$ denotes the number of times answer $a$ appears.
The solve rate quantifies answer reliability. High solve rates indicate reliable reference answers due to solution convergence, while low solve rates suggest inconsistent solutions that may indicate flawed problem formulations.

\subsection{Scoring}
\label{sec:scoring}
We evaluate the quality of generated problems and solutions with different scoring functions. The teacher's problems are scored based on \emph{difficulty} and \emph{diversity}, while the student receives scores for \emph{correctness}. Additionally,  both roles incorporate format scores to ensure parseable outputs.

\subsubsection{Teacher Scoring}
\label{sec:teacher-scoring}
We capture novelty through two fundamental dimensions: difficulty and diversity. We measure difficulty using \emph{solvability} to ensure problems remain appropriately challenging and \emph{solution length} to encourage multi-step reasoning, as these provide complementary signals about problem difficulty. Diversity is promoted through embedding distance, which encourages exploration of varied mathematical concepts. These components form a unified novelty score that guides problem generation.

\paragraph{Solvability ($\text{score}_{\text{sol}}$).}
The solvability score identifies problems with appropriate challenge. We use solve rate as a proxy for solvability---problems with $s_{q_i} > s_{\max}$ are likely too easy, while those with $s_{q_i} < s_{\min}$ are either too difficult or malformed. We employ a triangular scoring function that peaks at the optimal solve rate and decreases linearly as problems become too easy or too hard.

We define the solve rate range as $[s_{\min}, s_{\max}]$. Easy problems ($s_{q_i} > s_{\max}$) fail to challenge the model, while problems that are too hard or malformed ($s_{q_i} < s_{\min}$) offer minimal training value.

Formally, for $s_{q_i} \in [0, 1]$, let $s_{\text{mid}} = (s_{\min}+s_{\max})/2$ be the midpoint:
\begin{equation}
  \text{score}_{\text{sol}}(q_i)=
  \begin{cases}
    1-\eta |s_{q_i}-s_{\text{mid}}| & \text{if } s_{q_i} \in[s_{\min},s_{\max}], \\
    0                               & \text{otherwise}
  \end{cases}
\end{equation}
where $\eta = \left (1 - 1/G \right ) / (s_{\text{mid}} - s_{\min})$ is the slope coefficient, with $G$ being the number of solution attempts. The score peaks at the midpoint $s_{\text{mid}}$ and decreases to $1/G$ at the boundaries.

This creates a symmetric triangular score centred at the midpoint of the solve rate range, giving a maximum score for problems with moderate difficulty and progressively less score as the solve rate approaches either boundary.

\paragraph{Solution Length ($\text{score}_{\text{len}}$).}
Solution length complements solvability by measuring problem complexity.
Problems requiring multi-step reasoning typically elicit longer solutions. We score problems using the average length of student solutions:
\begin{equation}
  \text{score}_{\text{len}}(q_i) = \min\left(\frac{\bar{l}(q_i)}{l_{\text{base}}}, \frac{l_{\text{cap}}}{l_{\text{base}}}\right)
\end{equation}
where $\bar{l}(q_i)$ denotes average solution length for problem $q_i$, $l_{\text{base}}$ is a normalisation factor (defaults to 1000 tokens), and $l_{\text{cap}}$ prevents outliers from dominating the scoring signal. This score complements the solvability score (see Appendix~\ref{sec:sol-len}).

\paragraph{Diversity ($\text{score}_{\text{div}}$).} We compute the semantic distance between each new problem and the existing problem pool:
\begin{equation}
  \text{score}_{\text{div}}(q_i) = \min_{q' \in \mathcal{P}_{t-1}} d(e_{q_i}, e_{q'})
\end{equation}
where $e_{q_i}$ and $e_{q'}$ represent problem embeddings obtained from a pre-trained encoder, and $d(\cdot, \cdot)$ denotes cosine distance. This score maximises when a problem is semantically distant from all existing problems in the pool.

\paragraph{Format ($\text{score}_{\text{fom}}^T$).}
The format score ensures proper problem structure. Generated problems must be enclosed in \textless question\textgreater{} tags with concepts listed in \textless concepts\textgreater{} tags (maximum three concepts). We assign $\text{score}_{\text{fom}}^T(q_i) = 1$ for correct formatting and $\text{score}_{\text{fom}}^T(q_i) = 0$ otherwise.

\paragraph{Novelty Score.} We combine these components into a novelty score capturing both difficulty and diversity:
\begin{equation}\label{eq:novelty}
  \begin{split}
    \text{score}_{\text{novel}}(q_i) = {} & \alpha \text{score}_{\text{sol}}(q_i) + \lambda
    \text{score}_{\text{len}}(q_i)                                                                                           + \gamma \text{score}_{\text{div}}(q_i) + \delta \text{score}_{\text{fom}}^T(q_i)
  \end{split}
\end{equation}
where $\alpha$, $\lambda$, $\gamma$, $\delta$ are hyperparameters that control the relative importance of each component. This novelty score is used to select high-quality problem-solution pairs for training.

\subsubsection{Student Scoring}
\label{sec:student-scoring}
The student's score is based on solution correctness. For each solution attempt, we evaluate correctness by comparing the parsed answer against the reference answer from majority voting.

\paragraph{Format ($\text{score}_{\text{fom}}^S$).}
The format score ensures proper answer presentation. Solutions must present final answers in $\backslash$boxed\{\} notation. We assign $\text{score}_{\text{fom}}^S(o_j) = 1$ for correct formatting and $0$ otherwise.

\paragraph{Correctness Score.}
The student's correctness score combines accuracy with the format score:
\begin{equation}\label{eq:correct}
  \text{score}_{\text{correct}}(o_j, a_j) = \mathbf{1}[a_j = a^*] + \delta \text{score}_{\text{fom}}^S(o_j)
\end{equation}
where $\mathbf{1}[a_j = a^*]$ is an indicator function that equals 1 when parsed answer $a_j$ from outcome $o_j$ matches the reference answer $a^*$, and 0 otherwise. This correctness score evaluates both solution accuracy and proper formatting.

\subsection{Model Update}
\label{sec:problem-selection}

After computing novelty scores, we select $B$ high-quality samples from valid problems for reinforcement learning, allocating half to problem generation and half to solution solving.
For teacher training, we choose problem groups with highest $\text{score}_{\text{novel}}$ variance to ensure diverse training signals. For student training, we select problems with the highest novelty scores to provide maximal training value.

We optimise the policy using $\pi_\theta$ with an objective similar to Group Relative Policy Optimisation (GRPO) \citep{shao_deepseekmath_2024}, adapted for on-policy training to ensure stability \citep{chen_acereason-nemotron_2025}:
\begin{equation}
  \begin{split}
    \mathcal{J}(\theta) = {} & \mathbb{E}_{\substack{q_{1:G} \sim \pi_\theta(\cdot|p_T) \\ o_{1:G} \sim
      \pi_\theta(\cdot|q_i, p_S)}}\left[\sum_{r \in \{T, S\}} \frac{1}{G} \sum_{i=1}^G A_i^r\right]       - \beta \mathbb{D}_{\mathrm{KL}}\left(\pi_\theta \| \pi_{\mathrm{ref}}\right)
  \end{split}
\end{equation}
where $p_T$ and $p_S$ are the teacher and student prompts respectively, $r \in \{T, S\}$ refers to teacher and student, $\mathbb{D}_{KL}$ denotes the KL divergence, $\pi_{\mathrm{ref}}$ refers to the initial model before training.
The advantage for each role $r \in \{T, S\}$ is computed as:
\begin{equation}
  A_i^r = \frac{R_i^r - \operatorname{mean}\left(R_{1:G}^r\right)}{\operatorname{std}\left(R_{1:G}^r\right)}.
\end{equation}
We define role-specific rewards $R_i^T$ and $R_j^S$ using the scoring functions from Section \ref{sec:scoring}:
\begin{equation}
  R_i^T = \text{score}_{\text{novel}}(q_i), \quad R_j^S = \text{score}_{\text{correct}}(o_j, a_j)
\end{equation}
All valid problems are then added to the problem pool $\mathcal{P}_t$ for future iterations.
The full pseudocode is provided in Algorithm~\ref{alg:opensir} (Appendix~\ref{sec:algorithm}).

\section{Experiments}
\label{sec:experiments}
\subsection{Experimental Setup}

\paragraph{Training Details.}
We experiment with two instruction-tuned models: Llama-3.2-3B-Instruct and Llama-3.1-8B-Instruct~\citep{dubey2024llama} with GRPO \citep{shao_deepseekmath_2024}, and two reasoning models: DeepSeek-R1-Distill-Llama-8B~\citep{deepseek-ai_deepseek-r1_2025} and Qwen3-8B~\citep{qwen3_2025}.
We use a learning rate of $3 \times 10^{-7}$, 20 warm-up steps, and KL divergence coefficient $10^{-4}$.
All methods are trained on an equivalent number of problem-solution pairs.
Each experiment is run with three random seeds.
Full training and evaluation details are in Appendix~\ref{sec:train_details} and~\ref{sec:eval_details}.

\paragraph{Datasets.}
We evaluate on seven mathematical benchmarks: GSM8K \citep{cobbe_training_2021}, MATH-500 \citep{hendrycks_measuring_2021}, Minerva \citep{lewkowycz_solving_2022}, OlympiadBench \citep{he_olympiadbench_2024}, College Math \citep{tang_mathscale_2024}, and AIME 2024 and 2025; and three general reasoning benchmarks: BBEH \citep{kazemi_big-bench_2025}, MMLU-Pro \citep{wang_mmlu-pro_2024}, and SuperGPQA \citep{du_supergpqa_2025}.
We report Pass@1, averaged over 16 independently sampled generations per problem.

\paragraph{Baselines.}
(1) \textbf{Base}: zero-shot prompting with step-by-step reasoning.
(2) \textbf{GRPO$_\text{gsm8k}$} and \textbf{GRPO$_\text{math}$}: GRPO \citep{shao_deepseekmath_2024} trained on GSM8K (7,473 examples) \citep{cobbe_training_2021} and MATH (7,500 examples) \citep{hendrycks_measuring_2021}.
(3) \textbf{Absolute Zero} \citep{zhao_absolute_2025}: self-play using Python as external verifier.
(4) \textbf{R-Zero} \citep{huang_r-zero_2025}: verifier-free self-play with separate challenger and solver models using repetition penalties.

\subsection{Main Results}
\label{sec:main-results}

\begin{table*}[!th]
  \centering
  \footnotesize
  \setlength{\tabcolsep}{1.2pt}
  \begin{tabular}{l|ccccccc|c|ccc|c|c}
    \toprule
    \multirow{3}{*}{\textbf{Method}}  & \multicolumn{8}{|c|}{\textbf{Mathematical Reasoning}} & \multicolumn{4}{c|}{\textbf{General Reasoning}} & \multirow{3}{*}{\shortstack[c]{\textbf{Overall}                                                                                                                                                                                                                                              \\\textbf{Avg.}}} \\
    \cmidrule(lr){2-9} \cmidrule(lr){10-13}
                                      & \multirow{2}{*}{\textbf{GSM8K}}                       & \textbf{MATH}                                   & \multirow{2}{*}{\textbf{Minerva}}               & \textbf{College} & \multirow{2}{*}{\textbf{Olymp.}} & \textbf{AIME} & \textbf{AIME} & \multirow{2}{*}{\textbf{Avg.}} & \multirow{2}{*}{\textbf{BBEH}} & \textbf{MMLU-} & \textbf{Super-} & \multirow{2}{*}{\textbf{Avg.}} &                \\
                                      &                                                       & \textbf{500}                                    &                                                 & \textbf{Math}    &                                  & \textbf{24}   & \textbf{25}   &                                &                                & \textbf{Pro}   & \textbf{GPQA}   &                                &                \\
    \midrule
    \multicolumn{14}{l}{\textbf{Llama-3.2-3B-Instruct}}                                                                                                                                                                                                                                                                                                                                                                                        \\
    Base                              & 73.94                                                 & 42.86                                           & 15.21                                           & 28.78            & 13.09                            & 5.00          & 0.21          & 25.58                          & 4.49                           & 24.91          & 13.54           & 14.31                          & 19.95          \\
    GRPO$_{\text{gsm8k}}$             & \textbf{79.72}                                        & 45.30                                           & 16.27                                           & 33.33            & 14.56                            & 6.82          & 1.88          & 28.27                          & 6.12                           & 25.38          & 15.63           & 15.71                          & 21.99          \\
    GRPO$_{\text{math}}$              & 76.48                                                 & 45.26                                           & 16.09                                           & 32.95            & 14.13                            & 6.47          & 0.92          & 27.47                          & 5.98                           & 25.59          & 14.87           & 15.48                          & 21.48          \\
    Absolute Zero                     & 74.37                                                 & 44.71                                           & 14.78                                           & 31.93            & 14.42                            & 5.53          & 1.05          & 26.68                          & 4.54                           & 25.16          & 12.64           & 14.11                          & 20.40          \\
    R-Zero                            & 76.34                                                 & 44.27                                           & 15.84                                           & 32.72            & 14.19                            & 5.71          & 1.13          & 27.17                          & 4.73                           & 25.21          & 13.48           & 14.47                          & 20.82          \\
    \rowcolor{lightblue!50} {\method} & 78.28                                                 & \textbf{46.22}                                  & \textbf{18.46}                                  & \textbf{35.42}   & \textbf{16.72}                   & \textbf{7.94} & \textbf{3.97} & \textbf{29.57}                 & \textbf{7.13}                  & \textbf{27.32} & \textbf{16.60}  & \textbf{17.02}                 & \textbf{23.30} \\
    \midrule
    \multicolumn{14}{l}{\textbf{Llama-3.1-8B-Instruct}}                                                                                                                                                                                                                                                                                                                                                                                        \\
    Base                              & 84.50                                                 & 47.89                                           & 22.75                                           & 34.10            & 16.26                            & 3.12          & 0.42          & 29.86                          & 6.48                           & 32.85          & 16.57           & 18.63                          & 24.25          \\
    GRPO$_{\text{gsm8k}}$             & \textbf{88.70}                                        & 50.37                                           & 24.83                                           & 35.03            & 16.43                            & 5.24          & 1.78          & 31.77                          & 7.93                           & 34.18          & 17.77           & 19.96                          & 25.87          \\
    GRPO$_{\text{math}}$              & 86.23                                                 & 50.82                                           & 23.98                                           & 34.93            & 16.54                            & 4.39          & 1.92          & 31.26                          & 7.79                           & 33.27          & 16.65           & 19.24                          & 25.25          \\
    Absolute Zero                     & 86.89                                                 & 51.38                                           & 23.21                                           & 34.39            & 15.96                            & 2.78          & 0.65          & 30.75                          & 6.76                           & 33.08          & 16.66           & 18.83                          & 24.79          \\
    R-Zero                            & 86.19                                                 & 50.93                                           & 24.11                                           & 32.93            & 15.66                            & 2.63          & 1.08          & 30.50                          & 6.63                           & 32.92          & 16.63           & 18.73                          & 24.62          \\
    \rowcolor{lightblue!50} {\method} & 87.30                                                 & \textbf{52.38}                                  & \textbf{27.29}                                  & \textbf{36.29}   & \textbf{17.81}                   & \textbf{7.93} & \textbf{2.94} & \textbf{33.13}                 & \textbf{9.94}                  & \textbf{36.48} & \textbf{20.16}  & \textbf{22.19}                 & \textbf{27.66} \\
    \bottomrule
  \end{tabular}
  \caption{Results on instruction-tuned models (Pass@1). {\method} outperforms GRPO baselines trained on $>$7,000 human-annotated examples and self-play methods across model families, starting from a single seed problem.}
  \label{tab:math-results}
\end{table*}

\textbf{Instruction models.}\quad Table~\ref{tab:math-results} shows that
Absolute Zero and R-Zero yield at most $+$1.59 on Llama-3.2-3B-Instruct
and $+$0.89 on Llama-3.1-8B-Instruct in average math performance, trailing the GRPO
baselines despite much larger improvements on base models in their original
work~\citep{zhao_absolute_2025,huang_r-zero_2025}, suggesting that self-play
without explicit novelty incentives provides limited additional benefit.
  {\method} achieves the largest gains on the 3B and 8B models ($+$3.99, $+$3.27),
surpassing both GRPO$_{\text{gsm8k}}$ ($+$2.69, $+$1.91) and
GRPO$_{\text{math}}$ ($+$1.89, $+$1.40).
It does so from a single seed problem, whereas both GRPO baselines
require more than 7,000 human-annotated examples.
This indicates that problem diversity is critical for effective self-play on
instruction-tuned models; we examine this further in our diversity ablation
(\S\ref{sec:diversity}).

\begin{table*}[!th]
  \centering
  \footnotesize
  \setlength{\tabcolsep}{1.2pt}
  \begin{tabular}{l|ccccccc|c|ccc|c|c}
    \toprule
    \multirow{3}{*}{\textbf{Method}}  & \multicolumn{8}{|c|}{\textbf{Mathematical Reasoning}} & \multicolumn{4}{c|}{\textbf{General Reasoning}} & \multirow{3}{*}{\shortstack[c]{\textbf{Overall}                                                                                                                                                                                                                                                \\\textbf{Avg.}}} \\
    \cmidrule(lr){2-9} \cmidrule(lr){10-13}
                                      & \multirow{2}{*}{\textbf{GSM8K}}                       & \textbf{MATH}                                   & \multirow{2}{*}{\textbf{Minerva}}               & \textbf{College} & \multirow{2}{*}{\textbf{Olymp.}} & \textbf{AIME}  & \textbf{AIME}  & \multirow{2}{*}{\textbf{Avg.}} & \multirow{2}{*}{\textbf{BBEH}} & \textbf{MMLU-} & \textbf{Super-} & \multirow{2}{*}{\textbf{Avg.}} &                \\
                                      &                                                       & \textbf{500}                                    &                                                 & \textbf{Math}    &                                  & \textbf{24}    & \textbf{25}    &                                &                                & \textbf{Pro}   & \textbf{GPQA}   &                                &                \\
    \midrule
    \multicolumn{14}{l}{\textbf{DeepSeek-R1-Distill-Llama-8B}}                                                                                                                                                                                                                                                                                                                                                                                   \\
    Base                              & 70.68                                                 & 80.75                                           & 32.40                                           & 48.27            & 55.75                            & 41.04          & 28.12          & 51.00                          & 7.15                           & 40.33          & 17.38           & 21.62                          & 36.31          \\
    GRPO$_{\text{gsm8k}}$             & 72.49                                                 & 79.94                                           & 32.42                                           & 49.38            & 57.29                            & 40.38          & 28.83          & 51.53                          & 7.17                           & 41.35          & 18.72           & 22.41                          & 36.97          \\
    GRPO$_{\text{math}}$              & 71.38                                                 & 81.89                                           & 33.12                                           & 50.29            & 57.84                            & 39.41          & 29.24          & 51.88                          & 8.32                           & 40.42          & 19.24           & 22.66                          & 37.27          \\
    Absolute Zero                     & 71.93                                                 & 80.83                                           & 29.49                                           & 48.71            & 56.42                            & 39.72          & 27.94          & 50.72                          & 6.04                           & 39.37          & 16.40           & 20.60                          & 35.66          \\
    R-Zero                            & 73.78                                                 & 80.44                                           & 31.29                                           & 49.83            & 55.39                            & 40.87          & 28.42          & 51.43                          & 7.29                           & 40.45          & 15.45           & 21.06                          & 36.25          \\
    \rowcolor{lightblue!50} {\method} & \textbf{78.39}                                        & \textbf{84.68}                                  & \textbf{35.88}                                  & \textbf{51.42}   & \textbf{57.92}                   & \textbf{43.38} & \textbf{31.29} & \textbf{54.71}                 & \textbf{11.45}                 & \textbf{44.18} & \textbf{23.59}  & \textbf{26.41}                 & \textbf{40.56} \\
    \midrule
    \multicolumn{14}{l}{\textbf{Qwen3-8B}}                                                                                                                                                                                                                                                                                                                                                                                                       \\
    Base                              & 95.93                                                 & 97.02                                           & 48.44                                           & 58.45            & 75.18                            & 75.21          & 66.25          & 73.78                          & 15.35                          & 64.57          & 34.06           & 37.99                          & 55.89          \\
    GRPO$_{\text{gsm8k}}$             & \textbf{96.84}                                        & 96.94                                           & 49.12                                           & 57.94            & 73.38                            & 74.78          & 64.29          & 73.33                          & 14.31                          & 62.53          & 34.03           & 36.96                          & 55.15          \\
    GRPO$_{\text{math}}$              & 96.25                                                 & \textbf{98.32}                                  & 48.43                                           & 59.29            & 74.57                            & 77.84          & 65.13          & 74.26                          & 13.49                          & 65.34          & 35.43           & 38.09                          & 56.18          \\
    Absolute Zero                     & 93.73                                                 & 94.14                                           & 48.71                                           & 57.32            & 72.29                            & 72.42          & 63.82          & 71.78                          & 15.41                          & 62.51          & 33.96           & 37.29                          & 54.53          \\
    R-Zero                            & 94.83                                                 & 93.83                                           & 49.89                                           & 52.38            & 71.48                            & 73.15          & 62.23          & 71.11                          & 12.53                          & 64.68          & 33.19           & 36.80                          & 53.96          \\
    \rowcolor{lightblue!50} {\method} & 96.34                                                 & 97.13                                           & \textbf{52.28}                                  & \textbf{61.37}   & \textbf{76.77}                   & \textbf{79.52} & \textbf{69.91} & \textbf{76.19}                 & \textbf{18.30}                 & \textbf{69.05} & \textbf{39.84}  & \textbf{42.40}                 & \textbf{59.29} \\
    \bottomrule
  \end{tabular}
  \caption{Results on reasoning models (Pass@1). All baselines show marginal or negative gains, while {\method} is the only method that consistently improves both mathematical and general reasoning.}
  \label{tab:thinking-results}
\end{table*}

\textbf{Reasoning models.}\quad\label{sec:thinking-models} We next examine models
with extensive reasoning-focused post-training, where less headroom remains for further gains.
As Table~\ref{tab:thinking-results} shows, {\method} improves
average math performance by $+$3.71 on DeepSeek-R1-Distill-Llama-8B and $+$2.41 on Qwen3-8B.
In contrast, the data-based GRPO baselines provide only small gains, with a best
result of $+$0.88, whereas the self-play baselines deteriorate in three of four
cases, demonstrating that {\method} can expand capabilities even after heavy
reasoning post-training and further supporting the value of novel problem generation.
We include training curves across all four models in Appendix~\ref{sec:training-curves}.

\textbf{General reasoning transfer.}\quad We now ask whether these gains extend beyond
mathematics. Across Tables~\ref{tab:math-results} and \ref{tab:thinking-results}, {\method} improves
general reasoning performance by $+$2.71 and $+$3.56 on Llama-3.2-3B-Instruct and
Llama-3.1-8B-Instruct respectively; gains are larger on reasoning models, reaching
$+$4.79 and $+$4.41 on DeepSeek-R1-Distill-Llama-8B and Qwen3-8B. The best baseline reaches only $+$1.40 on instruction models and is negative or negligible
on reasoning models, suggesting that training on diverse self-generated novel problems develops
transferable reasoning skills, not only mathematical proficiency. {\method} accordingly
achieves the highest overall average on all four base models ($+$3.35 to $+$4.25).

\subsection{Analysis}

We analyse each key component of {\method} on Llama-3.2-3B-Instruct.

\paragraph{Evolution of Difficulty and Diversity.}
\label{sec:evolution}
\begin{figure*}[htbp]
    \centering
    \includegraphics[width=.95\textwidth]{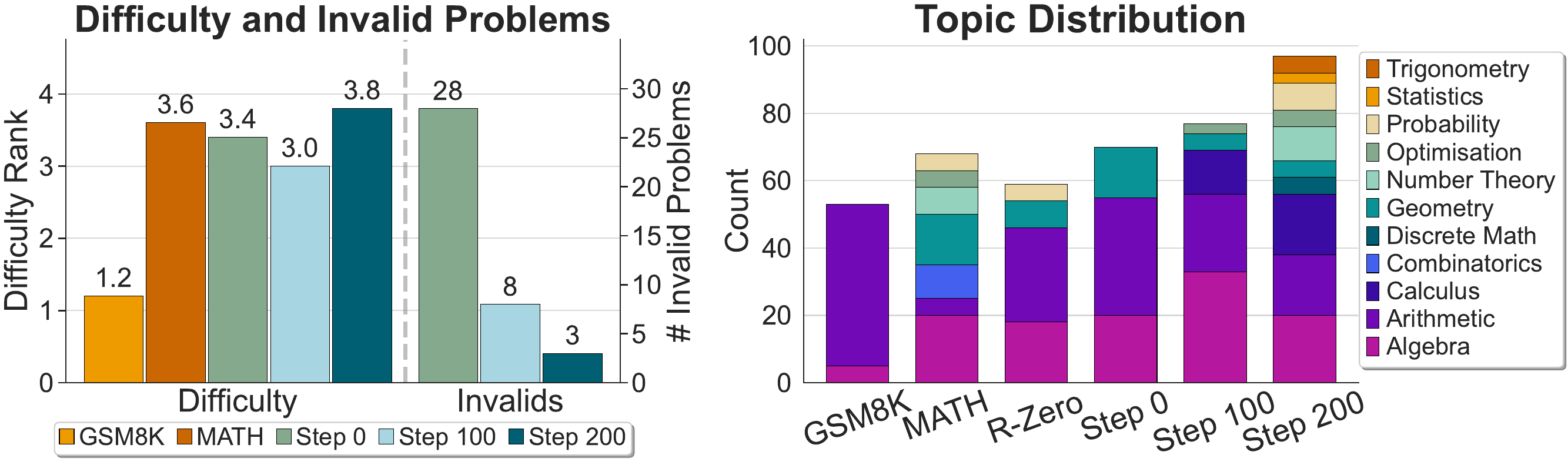}
    \caption{Evolution of problem difficulty, validity, and topic diversity during {\method} training (50 problems per source).
        \textbf{(Left)} Human evaluation results showing difficulty rankings (1--5 scale where 1=easiest, 5=hardest)
        and the number of invalid problems for GSM8K, MATH, and problems generated at steps 0, 100, and 200 of training.
        Invalid problems are those with logical flaws, missing information, or ambiguities.
        \textbf{(Right)} Distribution of mathematical topics across training stages and baselines.
        R-Zero's problems concentrate on a small number of topic categories, suggesting narrower coverage without an explicit diversity mechanism.
        In contrast, {\method} shows progressively broader topic coverage from step 0 to step 200.}
    \label{fig:annotation-results}
\end{figure*}

We track how difficulty and diversity evolve during training through human evaluation.
We sample 50 problems from three {\method} training checkpoints (steps 0, 100, 200) and 50 each from GSM8K and MATH. Annotators evaluate mixed sets of five problems (one per source), identifying topics, assessing validity, and ranking difficulty. We additionally sample 50 problems from R-Zero for topic annotation. Figure \ref{fig:annotation-results} shows average difficulty rankings (1=easiest, 5=hardest); see Appendix \ref{sec:annotation_details} for full annotation instructions.

Figure \ref{fig:annotation-results} (left) suggests a V-shaped difficulty trend across training stages. Problems start at 3.4 average difficulty ranking, drop to 3.0 at midpoint, then rise to 3.8.
This pattern reflects {\method}'s self-calibration: the model first generates overly difficult problems, then learns appropriate difficulty, and finally increases challenge as its solving capabilities improve.
The model also generates increasingly valid problems during training --- validity improves from below 50\% initially to 94\% (47 of 50 problems) by the end.

Figure \ref{fig:annotation-results} (right) shows topic diversity expansion across training. {\method} progresses from basic topics (algebra, arithmetic, geometry) to advanced domains including calculus and optimisation, eventually incorporating trigonometry, statistics, and other mathematical areas.
In contrast, R-Zero's problems remain concentrated in a small number of topic categories, suggesting that self-play without an explicit diversity mechanism tends to produce repetitive problem distributions.
Appendix \ref{sec:case-study} provides detailed case studies that illustrate this evolution.

\paragraph{Difficulty-Validity Trade-off.}
\label{sec:solve-rate}
We investigate the difficulty-validity trade-off by training {\method} variants with lower solve-rate thresholds of 0.1, 0.3, and 0.5, keeping the upper threshold at 0.9.
From each variant, we sample 300 problems and assess quality with GPT-5 \citep{openai_gpt-5_2025} using 8 responses per problem. We measure validity as the agreement between GPT-5's majority answer and our reference answer, and difficulty by GPT-5's solve rate.


\begin{wraptable}{r}{0.5\textwidth}
  \centering
  \vspace{-12pt}
  \footnotesize
  \begin{tabular}{l|c|c|c}
    \toprule
    \textbf{Model}    & \textbf{Math Avg.} & \textbf{Validity} & \textbf{Solve Rate} \\
    \midrule
    {\method}$_{0.5}$ & \textbf{29.57}     & 70.82             & 89.82               \\
    {\method}$_{0.3}$ & 27.81              & 52.32             & 81.38               \\
    {\method}$_{0.1}$ & 25.97              & 42.31             & 78.31               \\
    \bottomrule
  \end{tabular}
  \caption{Lower solve-rate thresholds produce harder but less valid problems, reducing overall performance.}
  \label{tab:sr-results}
  \vspace{-12pt}
\end{wraptable}
Table \ref{tab:sr-results} reveals a steep trade-off between difficulty and validity.
Lowering the threshold from 0.5 to 0.1 produces moderately harder problems
(GPT-5 solve rate: 89.82\% $\to$ 78.31\%), but validity collapses from 70.82\%
to 42.31\%, suggesting that low-solve-rate problems are more often invalid
than genuinely challenging.
Downstream math accuracy confirms that this trade-off is decisively unfavourable:
performance drops monotonically from 29.57 to 27.81 to 25.97 as the threshold
is relaxed from 0.5 to 0.3 to 0.1, as the increased difficulty cannot
compensate for training on a growing share of invalid problems.
This supports our selection of 0.5 as the lower threshold for the solvability reward.
While the residual $\sim$30\% reference-answer disagreement at $s_{\min}=0.5$ is non-trivial, {\method} still outperforms GRPO baselines trained on human-annotated examples (Table~\ref{tab:math-results}), indicating that this level of pseudo-label noise does not impair downstream performance.

Besides solve-rate thresholds, we find that rewarding longer solutions provides another mechanism for promoting problem complexity that encourage sophisticated multi-step problems (Appendix~\ref{sec:sol-len}).

\paragraph{Impact of Diversity Rewards.}
\label{sec:diversity}

\begin{wrapfigure}{r}{0.5\textwidth}
    \centering
    \vspace{-12pt}
    \includegraphics[width=\linewidth]{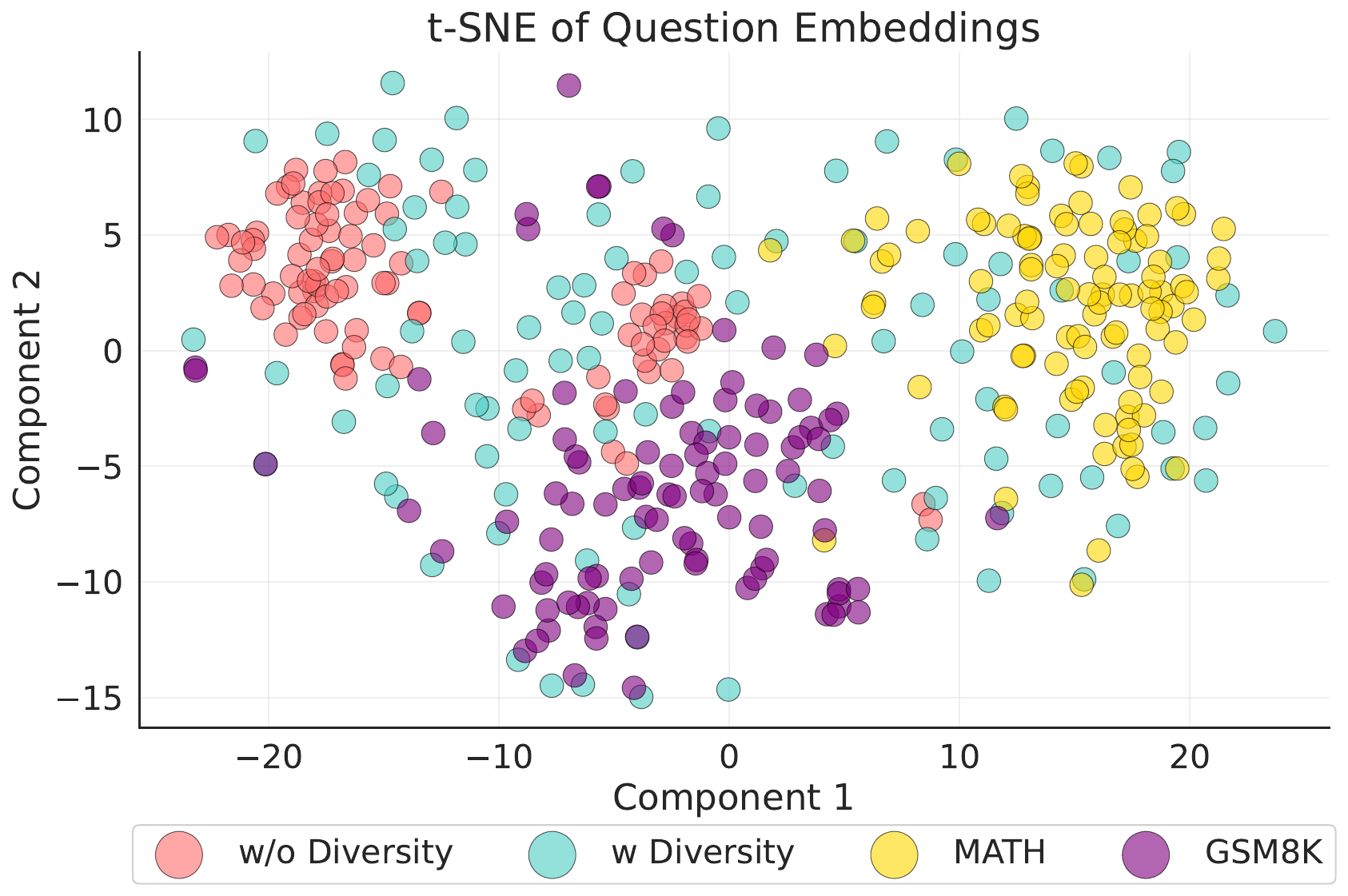}
    \captionof{figure}{Diversity reward spreads generated problems across a broader embedding space, while removing it yields a tighter cluster.}
    \label{fig:question_diversity}
    \vspace{-12pt}
\end{wrapfigure}

We analyse the impact of the diversity reward on problem diversity through problem embeddings, n-gram similarity, and concept overlap. Figure~\ref{fig:question_diversity} visualises the problem embeddings with t-SNE, where {\color{nodiversity}red} points represent problems without diversity reward, {\color{withdiversity}cyan} points show problems with diversity reward, {\color{mathcolor}gold} indicates MATH dataset problems, and {\color{gsmcolor}purple} marks GSM8K dataset problems. Without diversity rewards, problems cluster in narrow regions, generating similar types repeatedly and failing to achieve open-ended exploration. With diversity rewards, problems spread across the embedding space, reaching areas beyond MATH and GSM8K training sets. Further analysis of n-gram similarity and concept overlap support these findings, demonstrating consistent patterns of greater dispersion and novelty (Appendix~\ref{sec:ab_diversity}).

\begin{wraptable}{r}{0.5\textwidth}
    \centering
    \vspace{-12pt}
    \footnotesize
    \begin{tabular}{l|c|c|c}
        \toprule
        \multirow{2}{*}{\textbf{Model}} & \textbf{Math}  & \textbf{General} & \multirow{2}{*}{\textbf{\# Concepts}} \\
                                        & \textbf{Avg.}  & \textbf{Avg.}    &                                       \\
        \midrule
        w diversity                     & \textbf{29.57} & 17.02            & 5914                                  \\
        w/o diversity                   & 26.84          & 14.52            & 3328                                  \\
        \bottomrule
    \end{tabular}
    \captionof{table}{{\method} gains accuracy on math and general benchmarks with diversity reward, while nearly doubling concept coverage.}
    \label{tab:div-results}
    \vspace{-12pt}
\end{wraptable}

Table~\ref{tab:div-results} empirically confirms the importance of diversity rewards. Removing diversity rewards reduces mathematical reasoning performance by 2.73 (from 29.57 to 26.84) and general reasoning performance by 2.50 (from 17.02 to 14.52), alongside a sharp drop in the number of unique concepts explored (from 5914 to 3328). Importantly, the drop on general reasoning indicates that diversity rewards help the model develop more generalisable capabilities, rather than narrowly excelling on the mathematical distribution it was trained on.
This demonstrates that without diversity rewards, the model generates repetitive problems with limited learning value, constraining the teacher's ability to present varied mathematical challenges to the student.
Incorporating diversity rewards thus enables exploration of novel problems beyond existing datasets, supporting open-ended learning where the model continuously discovers new challenges rather than repeating known concepts.

\paragraph{Importance of Dual-Role Training.}
\label{sec:dual-role}
We evaluate the contribution of the joint teacher-student training by testing a variant where only the student is updated while the teacher remains fixed at its initial state.
Table~\ref{tab:teacher-results} shows that accuracy drops significantly from 29.57 to 25.89 when only the student is trained.
This suggests that effective self-play requires both components to co-evolve.

\begin{wraptable}{r}{0.5\textwidth}
  \centering
  \vspace{-12pt}
  \footnotesize
  \begin{tabular}{c|c|c|c}
    \toprule
    \textbf{Trained} & \textbf{Math}  & \textbf{General} & \textbf{Avg.}                      \\
    \textbf{Roles}   & \textbf{Avg.}  & \textbf{Avg.} & \textbf{Solve Rate}                \\
    \midrule
    Both             & \textbf{29.57} & \textbf{17.02}  & 72.20  {\footnotesize ($\pm$4.49)} \\
    Student          & 25.89          & 14.27 & 64.56 {\footnotesize ($\pm$17.37)} \\
    \bottomrule
  \end{tabular}
  \captionof{table}{Effect of teacher training. Joint training achieves higher accuracy and remarkably stable problem difficulty (much lower solve rate variance), demonstrating that teacher training enables calibrated problem generation at optimal difficulty levels for effective learning.}
  \label{tab:teacher-results}
  \vspace{-12pt}
\end{wraptable}

Without teacher training, generated problems become harder (solve rate drops from 72.20 to 64.56) and drift from the optimal 70\% target solve rate established in Section~\ref{sec:solve-rate}. More critically, solve rate variance increases tremendously (from $\pm$4.49 to $\pm$17.37), indicating highly inconsistent difficulty during training. This poorly calibrated curriculum explains the performance drop: the fixed teacher cannot adapt to the student's evolving capabilities, whereas joint training enables continuous difficulty calibration at the optimal challenge level.

\section{Related Work}
\paragraph{Self-play.}
Self-play achieved superhuman performance in games without human data, from
AlphaGo~\citep{silver_mastering_2016,silver_mastering_2017-1}, StarCraft II~\citep{vinyals_grandmaster_2019}, Poker~\citep{brown_superhuman_2019}, DotA~\citep{openai_dota_2019}, and Diplomacy~\citep{meta_fundamental_ai_research_diplomacy_team_fair_human-level_2022}.
\citet{baker_emergent_2019} show that agents can discover complex strategies with self-play, suggesting it is a promising avenue for continuous open-ended learning.
Recent works apply self-play to LLM reasoning: Absolute Zero \citep{zhao_absolute_2025}
and Spiral~\citep{liu_spiral_2025} rely on external verifiers or game rules that limit their use beyond specific domains. R-Zero~\citep{huang_r-zero_2025} attempts verifier-free self-play using repetition penalties rather than explicit exploration incentives; our topic analysis suggests this leads to narrower coverage than {\method}, constraining open-ended learning. In contrast, {\method} generates and solves problems without external supervision while actively promoting diversity to enable continuous discovery of novel mathematical concepts.

\paragraph{Reinforcement Learning with Verifiable Feedback (RLVF).}
RLVF drives recent advances in LLM reasoning \citep{openai_learning_2024,openai_introducing_2025,deepseek-ai_deepseek-r1_2025} but requires extensive human-annotated data for verifiable reward signals \citep{zeng_simplerl-zoo_2025}, creating scalability bottleneck and potentially limiting performance to human-level.
Recent works show that moderate-difficulty training samples provide optimal learning signals \citep{zheng_act_2025,sun_improving_2025}, while diverse problem types enhance mathematical reasoning  \citep{akter_nemotron-crossthink_2025,chen_acereason-nemotron_2025}. These insights directly motivate {\method} to optimise for appropriate difficulty calibration and diversity-driven exploration, enable models to learn math reasoning open-endedly without human supervision.

\section{Conclusions}
We present {\method}, a self-play framework that enables LLMs to autonomously learn to generate and solve novel problems without external supervision.
Starting from only a single trivial math problem, our framework outperforms GRPO-trained models that utilise thousands of human annotations across diverse model families. This approach demonstrates that models can effectively bootstrap mathematical reasoning through recursive self-improvement, eliminating dependence on extensive curated datasets.
Our analysis reveals that {\method} succeeds by combining difficulty calibration and diversity rewards to create an adaptive curriculum where models continuously discover and master increasingly challenging mathematical concepts.
Overall, {\method} represents a compelling paradigm for open-ended autonomous mathematical reasoning development, enabling models to recursively expand their capabilities beyond the boundaries of human-annotated data.

\paragraph{Limitations.}
Performance eventually plateaus after extended training, suggesting that sustaining novelty within one domain may require stronger or complementary exploration mechanisms.
Like prior self-play work, {\method} explores within a single domain (mathematics with verifiable answers); extending novelty-driven self-improvement across domains is a natural direction toward more open-ended learning.
The diversity reward might rely on the quality of the embedding model used for scoring; we do not ablate alternative embedding models, though the consistent improvements across all four base models suggest robustness of the overall approach.

\begin{ack}
  The authors would like to thank Aryo Pradipta Gema, Neel Rajani, Rohit Saxena (in alphabetical order) for the helpful discussions and feedback on the manuscript.
\end{ack}

\bibliographystyle{plainnat}
\bibliography{references,custom,pasquale}

\appendix
\FloatBarrier
\section{Extended Results and Analysis}


\subsection{Training Curves}
\label{sec:training-curves}
\begin{figure}[htbp]
  \centering
  \includegraphics[width=.8\textwidth]{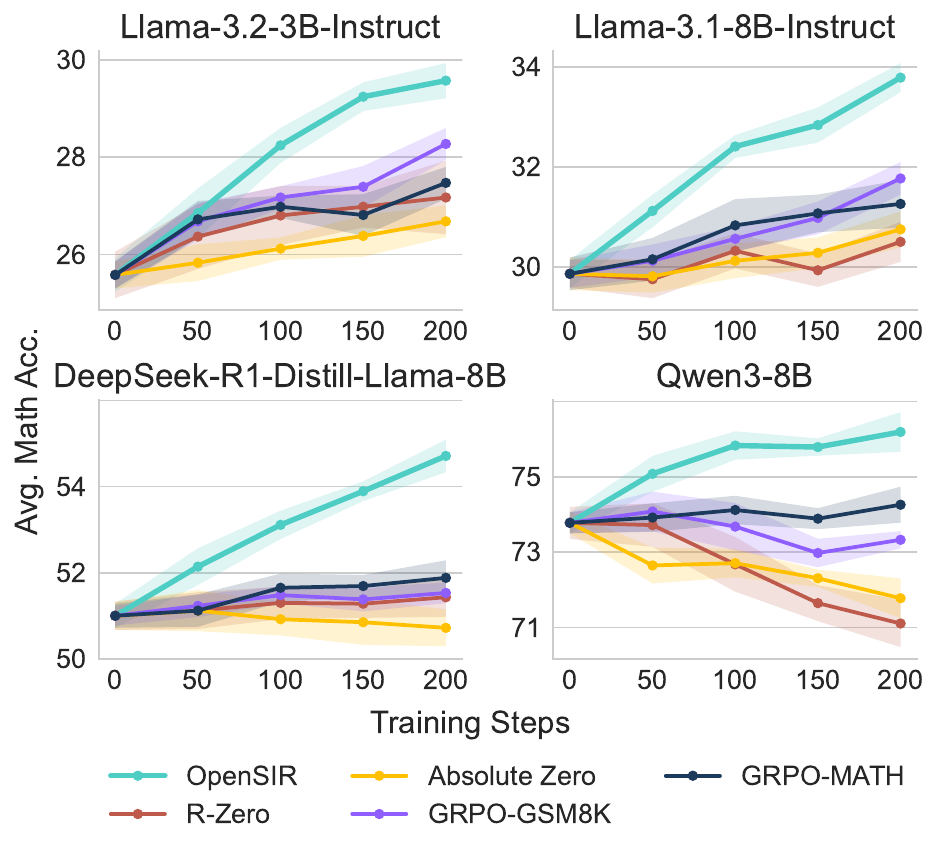}
  \caption{Average math accuracy over training steps for all four models.}
  \label{fig:training-curves}
\end{figure}

Figure~\ref{fig:training-curves} shows the average math accuracy over training steps for all four models. {\method} consistently improves throughout training, whereas baselines plateau or degrade.

\subsection{Case Study}
\label{sec:case-study}
This section provides further analysis of question-solution pairs during training.

As discussed in Section \ref{sec:evolution}, the model generates predominantly invalid problems early in training.
The majority of these problems, primarily involving simple mathematical concepts like arithmetic, fail due to missing information (Figures~\ref{fig:analysis_q1} and \ref{fig:analysis_q2}).
When attempting complex topics like optimisation, which are rare in the beginning, the model produces problems with missing information and fundamental formulation errors (Figure~\ref{fig:analysis_q3}). This reveals the model has a limited understanding of underlying mathematical concepts.
Invalid problems tend to exhibit low solve rates ($\leq$ 0.25) and correspondingly receive lower rewards, helping the model learn to generate valid problems. Consequently, invalid problems decrease rapidly across training (\S\ref{sec:evolution}).

However, not all problems with low solve rates are invalid (\S\ref{sec:solve-rate}).
We find that some problems involving certain topics that are challenging for the model, such as geometric series, persistently exhibit low solve rates (Figures~\ref{fig:analysis_q5}. The model struggles with exponentiation calculations, resulting in poor performance on geometric series problems.
This reveals a fundamental trade-off in {\method}: while higher solve rate thresholds effectively filter out invalid problems, they inevitably discourage exploration of genuinely difficult topics.
Since these problems have low solvability scores, they are likely to not receive sufficient encouragement to further explore these topics.

In later training stages, we observe {\method} gradually expanding into advanced mathematical domains. After 100 training steps, the model starts to generate problems involving concepts like optimisation (Figure~\ref{fig:analysis_q6}), calculus (Figure~\ref{fig:analysis_q7}), trigonometry-based physics (Figure~\ref{fig:analysis_q8}), probability (Figures~\ref{fig:analysis_q9}), among others.
While these advanced problems yield lower solve rates, which indicate the model has a limited understanding of these domains, they achieve high novelty scores with large semantic distances and longer solutions. This progression validates how novelty rewards in {\method} drive exploration of diverse mathematical concepts, enabling open-ended learning.

\begin{figure}[htbp]
  \centering
  \begin{tcolorbox}[colback=blue!5!white,colframe=blue!75!black,fontupper=\footnotesize,fontlower=\footnotesize,width=\textwidth ]
    A concert venue charges an admission price per seat and also offers a VIP ticket package that includes food, drinks, and other perks. If the food and other perks are included in the VIP ticket price and are worth \$10 per person, and a group of friends want to buy the same number of VIP tickets as they would with regular tickets, what is the minimum admission price per regular ticket to make the total cost of the VIP tickets equal to or less than the total cost of the regular tickets?
  \end{tcolorbox}
  \caption{An invalid arithmetic question generated in step 0 with solve rate of 0.25. This question is invalid since the VIP tick price is not provided, and therefore, it's impossible to calculate the minimum regular ticket price.}
  \label{fig:analysis_q1}
\end{figure}

\begin{figure}[htbp]
  \centering
  \begin{tcolorbox}[colback=blue!5!white,colframe=blue!75!black,fontupper=\footnotesize,fontlower=\footnotesize,width=\textwidth ]
    Find the percentage difference in the cumulative growth of two continuously compounded interest functions after 5 years: $A = P*e^{(rt)}$, where A is the amount of money accumulated after n years, including interest, P is the principal amount, r is the annual interest rate, and t is the time the money is invested for.
  \end{tcolorbox}
  \caption{An invalid arithmetic question generated in step 0 with solve rate of 0.125. This question is invalid since the two interest rates and principal amounts are not provided. Hence, it's impossible to calculate the percentage difference with just the general formula provided.}
  \label{fig:analysis_q2}
\end{figure}

\begin{figure}[htbp]
  \centering
  \begin{tcolorbox}[colback=blue!5!white,colframe=blue!75!black,fontupper=\footnotesize,fontlower=\footnotesize,width=\textwidth ]
    Consider two positive integers m and n (m $\geq$ n). Given a 2x2 matrix of numbers where each element is a non-negative integer, find the maximum value of the following expression: $ax^2 + by^2 + cxy$, where a, b, and c are constants, subject to the constraint that the sum of any two elements in each row and column of the matrix are equal. What is the maximum possible value of $ax^2 + by^2 + cxy$?
  \end{tcolorbox}
  \caption{An invalid optimisation question generated in step 0 with solve rate of 0.125. This question is invalid because there are missing information about the constants a, b, and c. There are also ambiguities in the question, such as the role of m and n in the problem. It also did not explain what the elements of the matrix represent. Lastly, it contains problem formulation errors, specifically failing to specify constraints that ensure bounded solutions, demonstrating insufficient understanding of optimization problem structure.}
  \label{fig:analysis_q3}
\end{figure}

\begin{figure}[htbp]
  \centering
  \begin{tcolorbox}[colback=blue!5!white,colframe=blue!75!black,fontupper=\footnotesize,width=\textwidth,title={\textbf{Question}}]
    A company is adding new employees to its workforce. Initially, the company has 500 employees and adds 10 new employees every quarter. If this trend continues for the next 5 years, and each quarter the company adds a fixed percentage increase in the number of new employees (starting with 10\%), how many new employees can the company be expected to add during the first 2 years?
  \end{tcolorbox}

  \vspace{0.5em}

  \begin{tcolorbox}[colback=gray!5!white,colframe=gray!75!black,fontupper=\footnotesize,width=\textwidth,title={\textbf{Solution} (Incorrect)}]
    To find the total number of new employees added in the first 2 years, we need to calculate the number of new employees added each quarter for the first 2 years and then sum them up.\\
    \\
    First, let's calculate the number of quarters in 2 years: 2 years * 4 quarters/year = 8 quarters.\\
    \\
    The company starts with 10 new employees and adds 10\% more each quarter. To find the total number of new employees added each quarter, we can use the formula for compound interest:\\
    \\
    $A = P * (1 + r)^n$\\
    \\
    where:\\
    A = total number of new employees after n quarters\\
    P = initial number of new employees (10)\\
    r = percentage increase per quarter (10\% or 0.1)\\
    n = number of quarters\\
    \\
    However, since we want to find the total number of new employees added each quarter, we need to find the difference between A and the initial number of new employees (10). We can rewrite the formula as:\\
    \\
    $A - P = P * (1 + r)^n$\\
    $A - 10 = 10 * (1 + 0.1)^n$\\
    $A = 10 * (1.1)^n$\\
    \\
    Now, we can calculate A for each quarter:\\
    \\
    Quarter 1 (n = 1): A = $10 * (1.1)^1 = 11$\\
    Quarter 2 (n = 2): A = $10 * (1.1)^2 = 12.1$\\
    Quarter 3 (n = 3): A = $10 * (1.1)^3 = 13.21$\\
    Quarter 4 (n = 4): A = $10 * (1.1)^4 = 14.41$\\
    Quarter 5 (n = 5): A = $10 * (1.1)^5 = 15.71$\\
    Quarter 6 (n = 6): A = $10 * (1.1)^6 = 17.31$\\
    Quarter 7 (n = 7): A = $10 * (1.1)^7 = 19.33$\\
    Quarter 8 (n = 8): A = $10 * (1.1)^8 = 21.59$\\
    \\
    Now, we can calculate the total number of new employees added in the first 2 years by summing up the number of new employees added each quarter:\\
    \\
    Total = 11 + 12.1 + 13.21 + 14.41 + 15.71 + 17.31 + 19.33 + 21.59\\
    Total = 119.66\\
    \\
    Since we can't have a fraction of an employee, we can round down to the nearest whole number. \\
    \\
    Therefore, the final answer is: $\boxed{119}$
  \end{tcolorbox}

  \caption{A valid arithmetic problem involving geometric series with a solve rate of 0.125. The solution requires either summing quarterly employee additions or applying the geometric series formula: $a \cdot \frac{r^n-1}{r-1}$. Llama-3.2-3B-Instruct struggles with exponentiation calculations, resulting in poor performance on geometric series problems.}
  \label{fig:analysis_q5}
\end{figure}

\begin{figure}[htbp]
  \centering
  \begin{tcolorbox}[colback=blue!5!white,colframe=blue!75!black,fontupper=\footnotesize,fontlower=\footnotesize,width=\textwidth]
    A farmer has 10 different types of vegetables to store in a warehouse with 1,000 square meters of floor space. There are four container sizes available:

    - Small (5 sq m): Maximum 50 available

    - Medium (10 sq m): Maximum 40 available

    - Large (15 sq m): Maximum 30 available

    - Extra-large (20 sq m): Maximum 25 available

    The vegetables have different storage requirements:

    - 3 bulky vegetables (pumpkins, watermelons, cabbages) require containers of at least 15 sq m

    - 4 medium vegetables (tomatoes, peppers, eggplants, zucchini) require containers of at least 10 sq m

    - 3 small vegetables (carrots, onions, potatoes) can fit in any container size

    Each vegetable type must be stored in at least one container. What is the maximum number of containers that can be used while satisfying all constraints and not exceeding 1,000 sq m total space?
  \end{tcolorbox}
  \caption{A valid optimisation problem with a solve rate of 0.375 generated at step 124.}
  \label{fig:analysis_q6}
\end{figure}

\begin{figure}[htbp]
  \centering
  \begin{tcolorbox}[colback=blue!5!white,colframe=blue!75!black,fontupper=\footnotesize,fontlower=\footnotesize,width=\textwidth]
    Find the equation of the curve y = f(x) where the derivative is given by f$^\prime$(x) = ($3x^2 - x - 2$)/2x and the curve passes through the point (2, 3).
  \end{tcolorbox}
  \caption{A valid calculus problem with a solve rate of 0.375 generated at step 156.}
  \label{fig:analysis_q7}
\end{figure}

\begin{figure}[htbp]
  \centering
  \begin{tcolorbox}[colback=blue!5!white,colframe=blue!75!black,fontupper=\footnotesize,fontlower=\footnotesize,width=\textwidth]
    A golfer hits a ball from the top of a 50-meter high cliff with an initial velocity of 30 m/s at an angle of 45 degrees above the horizontal. What is the horizontal distance traveled by the ball when it hits the ground?
  \end{tcolorbox}
  \caption{A valid physics problem that involves trigonometry with a solve rate of 0.5 generated at step 172.}
  \label{fig:analysis_q8}
\end{figure}

\begin{figure}[htbp]
  \centering
  \begin{tcolorbox}[colback=blue!5!white,colframe=blue!75!black,fontupper=\footnotesize,fontlower=\footnotesize,width=\textwidth]
    Consider a randomly ordered sequence of $n = 3q$ distinct integers $\{a_1, a_2, \ldots, a_{3q}\}$ where $q$ is a positive integer. Define $f$ as the number of adjacent pairs $(a_i, a_{i+1})$ in the sequence where both integers have the same remainder when divided by 3 (i.e., $a_i \bmod 3 = a_{i+1} \bmod 3$). If the integers $1$ through $3q$ are randomly permuted to form this sequence, what is the expected value of $f$?
  \end{tcolorbox}
  \caption{A valid probability problem with a solve rate of 0.25 generated at step 188.}
  \label{fig:analysis_q9}
\end{figure}

\subsection{Further Analysis on Questions Diversity}
\label{sec:ab_diversity}
\begin{figure}[htbp]
  \centering
  \includegraphics[width=\textwidth]{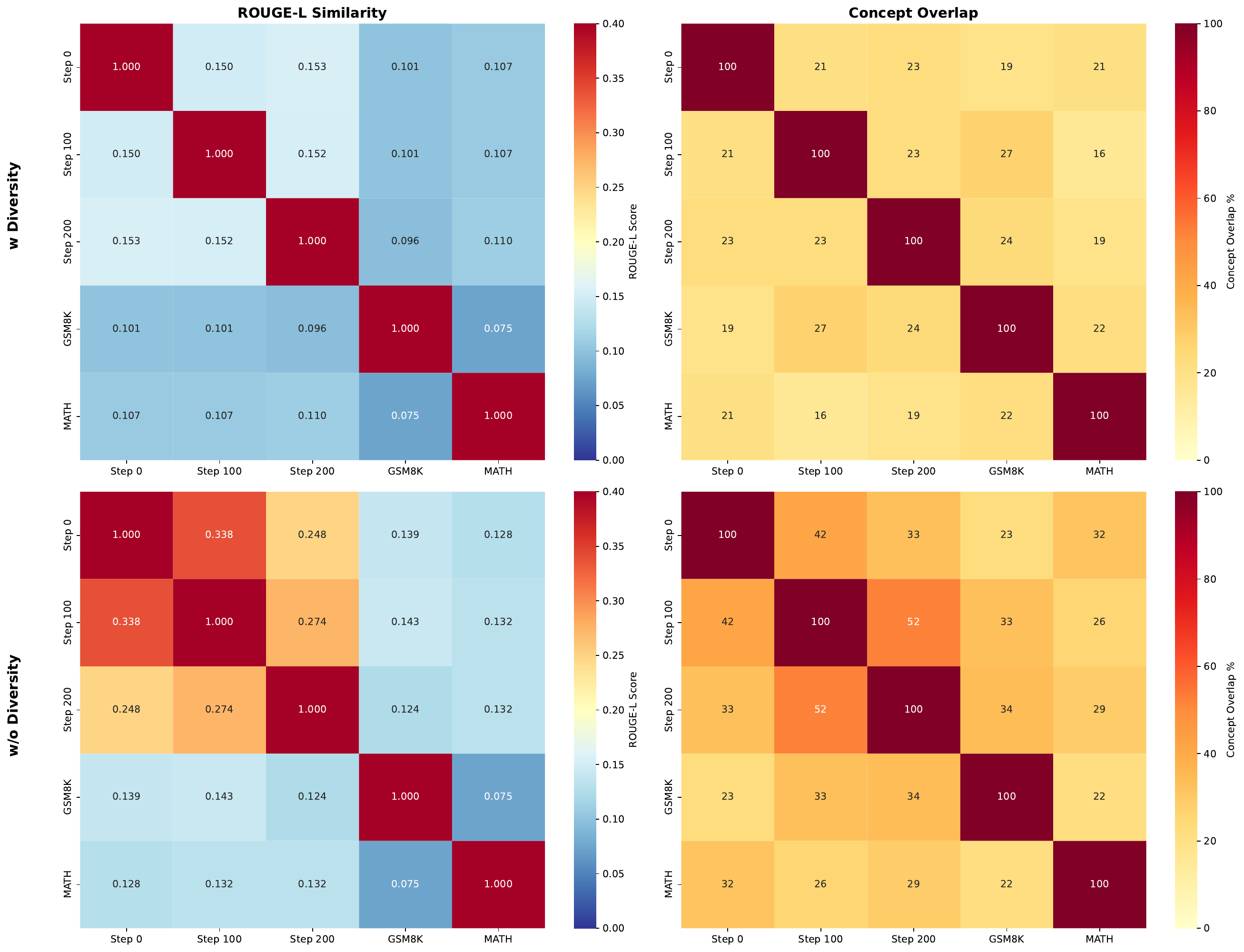}
  \caption{Heatmap visualisation of n-gram similarity (ROUGE-L scores) and concept overlap between generated problems at training steps 0, 100, 200 and reference datasets (MATH, GSM8K). Top row: with diversity reward; Bottom row: without diversity reward. With diversity reward incorporated, the generated problems exhibit low textual similarity and minimal concept overlap, demonstrating effective exploration of diverse problem types.}
  \label{fig:n-gram_diversity}
\end{figure}
Figure~\ref{fig:n-gram_diversity} presents n-gram similarity and concept analysis. We compute ROUGE-L scores between problem texts and extract mathematical concepts using GPT-5 from problems at steps 0, 100, and 200, as well as from the MATH and GSM8K training sets. With diversity rewards (top row), problems maintain low ROUGE-L scores and minimal concept overlap both across training stages and with MATH/GSM8K. Without diversity rewards (bottom row), both textual similarity and concept overlap increase, confirming limited exploration of new problem types.

\subsection{Further Analysis on Question Difficulty Progression}
\label{sec:appen_diversity}
\begin{figure}[htbp]
  \centering
  \includegraphics[width=.7\textwidth]{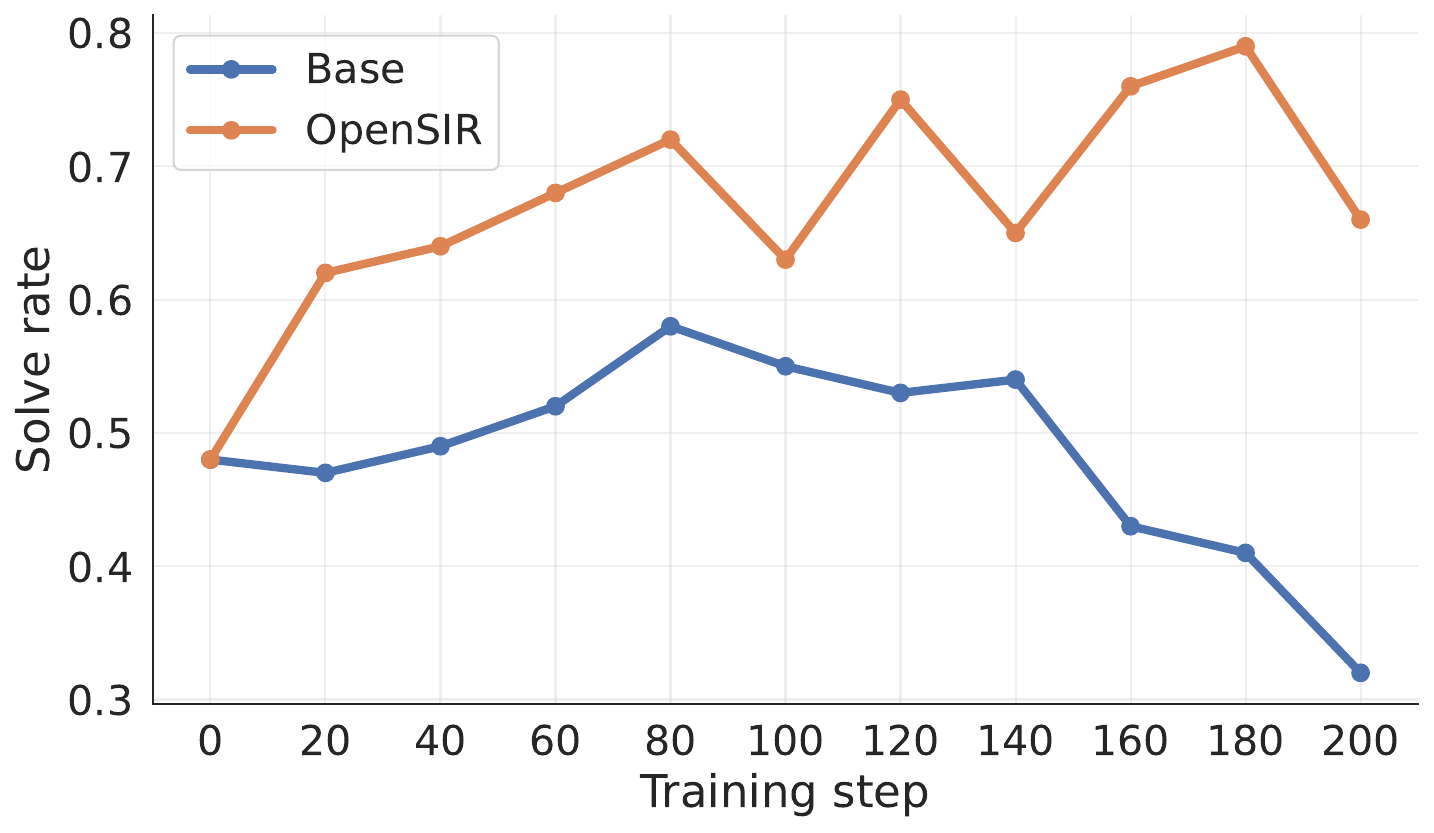}
  \caption{Progression of solve rates of {\method} and the initial instruction model as training goes.}
  \label{fig:solve_rates}
\end{figure}
Figure~\ref{fig:solve_rates} compares solve rates between the evolving {\method} policy and the fixed instruction model (Base) on problems generated during training.
While {\method}'s solve rate remains stable around 0.7 due to solvability-based problem selection (Section \ref{sec:teacher-scoring}), this approximately stable rate does not imply constant problem difficulty. As {\method} improves during training, maintaining the same solve rate requires generating progressively
harder problems. To verify this difficulty progression objectively, we measure how the initial instruction model (Base) performs on the same problems.
The base model's solve rate first rises (0.48$\to$0.58 at step 80) then declines (0.58$\to$0.32 at step 200), confirming the V-shaped difficulty pattern trend from Section \ref{sec:evolution}: problems initially become easier as {\method} learns appropriate calibration, then progressively harder as it increases challenge.
Crucially, this pattern shows that {\method}'s reasoning ability improved over training.

\subsection{Sensitivity to the initial seed problem}
To address whether {\method} can robustly escape the limited starting point of a trivial arithmetic seed (``What is 1+1?''), we experiment with two substantially different initial seeds using Llama-3.2-3B-Instruct: a geometry problem from the MATH dataset, representing a different mathematical domain, and a competition-level problem from AIME 2024, which is significantly more challenging than the trivial seed.

\begin{tcolorbox}[breakable,colback=blue!5!white,colframe=blue!75!black,fontupper=\footnotesize,fontlower=\small,width=\textwidth ]
  $B$ and $C$ trisect $\overline{AD}$ and $M$ is the midpoint of $\overline{AD}$. $MC = 8$. How many units are in the length of $\overline{AD}$?
\end{tcolorbox}
\captionof{figure}{A geometry problem from the MATH dataset, representing a different mathematical domain from the trivial arithmetic seed.}
\label{fig:math_seed_problem}

\begin{tcolorbox}[breakable,colback=blue!5!white,colframe=blue!75!black,fontupper=\footnotesize,fontlower=\small,width=\textwidth ]
  Every morning Aya goes for a $9$-kilometer-long walk and stops at a coffee shop afterwards. When she walks at a constant speed of $s$ kilometers per hour, the walk takes her 4 hours, including $t$ minutes spent in the coffee shop. When she walks $s+2$ kilometers per hour, the walk takes her 2 hours and 24 minutes, including $t$ minutes spent in the coffee shop. Suppose Aya walks at $s+\frac{1}{2}$ kilometers per hour. Find the number of minutes the walk takes her, including the $t$ minutes spent in the coffee shop.
\end{tcolorbox}
\captionof{figure}{A competition-level problem from AIME 2024, significantly more challenging than the trivial seed.}
\label{fig:aime_seed_problem}

\begin{table}[h]
  \centering
  \footnotesize
  \setlength{\tabcolsep}{3pt}
  \begin{tabular}{l|ccccccc|c}
    \toprule
    \multirow{2}{*}{\textbf{Model}} & \multirow{2}{*}{\textbf{GSM8K}} & \textbf{MATH} & \multirow{2}{*}{\textbf{Minerva}} & \textbf{College} & \multirow{2}{*}{\textbf{Olymp.}} & \textbf{AIME} & \textbf{AIME} & \textbf{Math} \\
                                    &                                 & \textbf{500}  &                                   & \textbf{Math}    &                                  & \textbf{24}   & \textbf{25}   & \textbf{Avg.} \\
    \midrule
    {\method}                       & 78.28                           & 46.22         & 18.46                             & 35.42            & 16.72                            & 7.94          & 3.97          & 29.57         \\
    {\method}$_{\text{MATH}}$       & 77.96                           & 46.48         & 18.72                             & 35.68            & 16.92                            & 7.82          & 4.22          & 29.72         \\
    {\method}$_{\text{AIME}}$       & 78.00                           & 45.90         & 18.20                             & 35.10            & 16.40                            & 8.20          & 3.86          & 29.38         \\
    \bottomrule
  \end{tabular}
  \caption{Performance of {\method} with different initial seed problem.}
  \label{tab:initial-seed-results}
\end{table}

Table \ref{tab:initial-seed-results} shows that all three variants achieve nearly identical performance (29.57, 29.72, and 29.38), with differences of less than 0.4 percentage points.
These results indicate that {\method} is robust to the initial seed problem, successfully escaping the limited starting point regardless of whether it begins with trivial arithmetic, a different mathematical domain (geometry), or a significantly more challenging competition-level problem.

\begin{figure}[htbp]
  \centering
  \includegraphics[width=.7\textwidth]{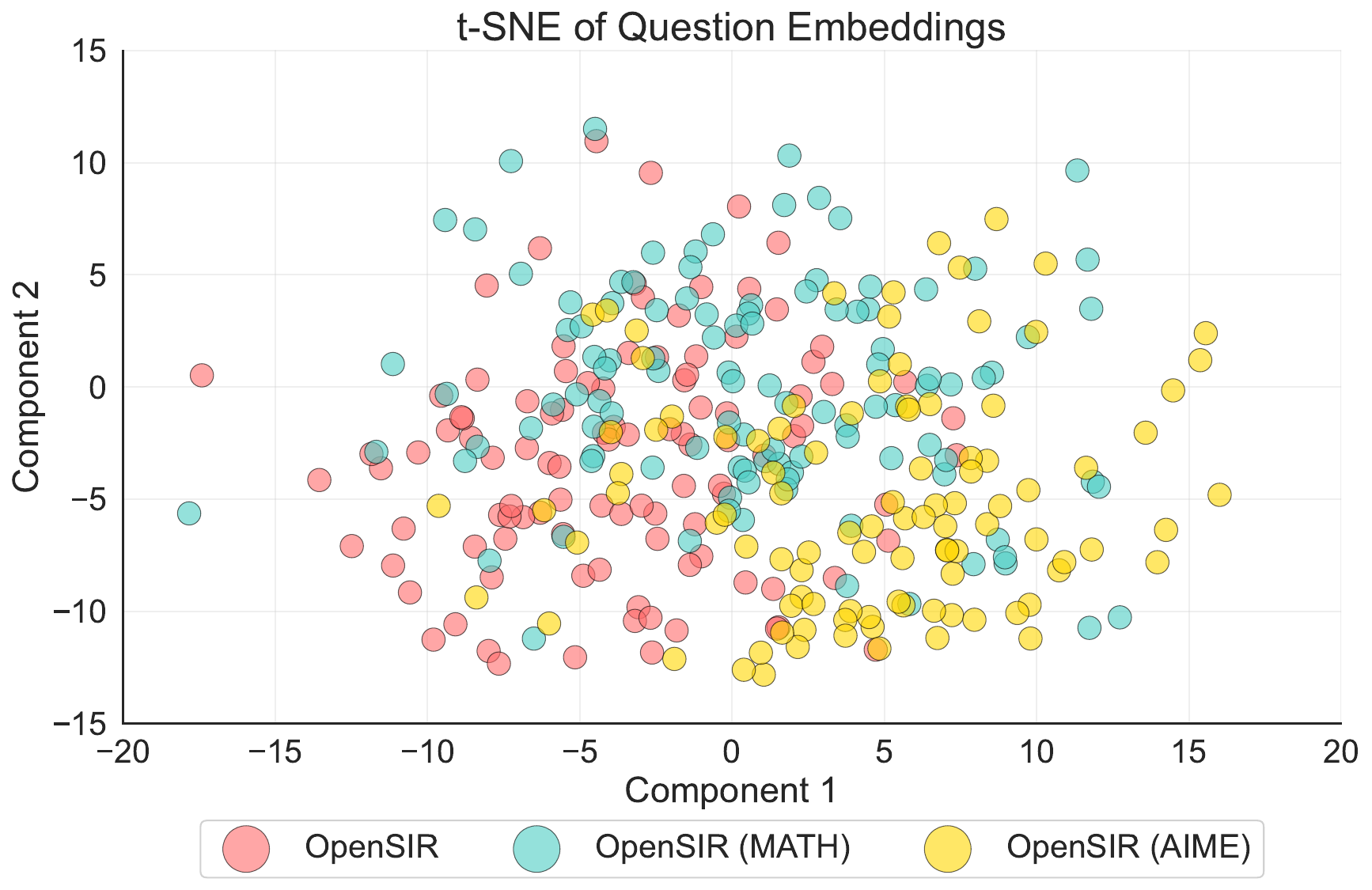}
  \caption{t-SNE visualisation of problem embeddings generated by {\method} from three different initial seeds. The substantial overlap demonstrates that the method converges to similar problem distributions regardless of the starting point.}
  \label{fig:different_seed_diversity}
\end{figure}

Figure \ref{fig:different_seed_diversity} visualises the diversity of problems generated at the final training step across the three different initial seeds.
The t-SNE embeddings reveal that all three variants produce diverse problems spanning similar regions of the semantic space, with substantial overlap in their distributions regardless of the initial seed.
This confirms that {\method} successfully escapes its starting point by exploring a wide range of mathematical concepts, driven by the diversity and solvability rewards that encourage continuous exploration beyond the initial problem domain and difficulty level.
\FloatBarrier
\subsection{{\method} Incentivises Reasoning Capacity}
To investigate whether {\method} elicits reasoning improvements rather than memorisation, we evaluate pass@k performance on five challenging mathematical benchmarks following \citep{yue_does_2025}.

\begin{figure}[htbp]
  \centering
  \includegraphics[width=\textwidth]{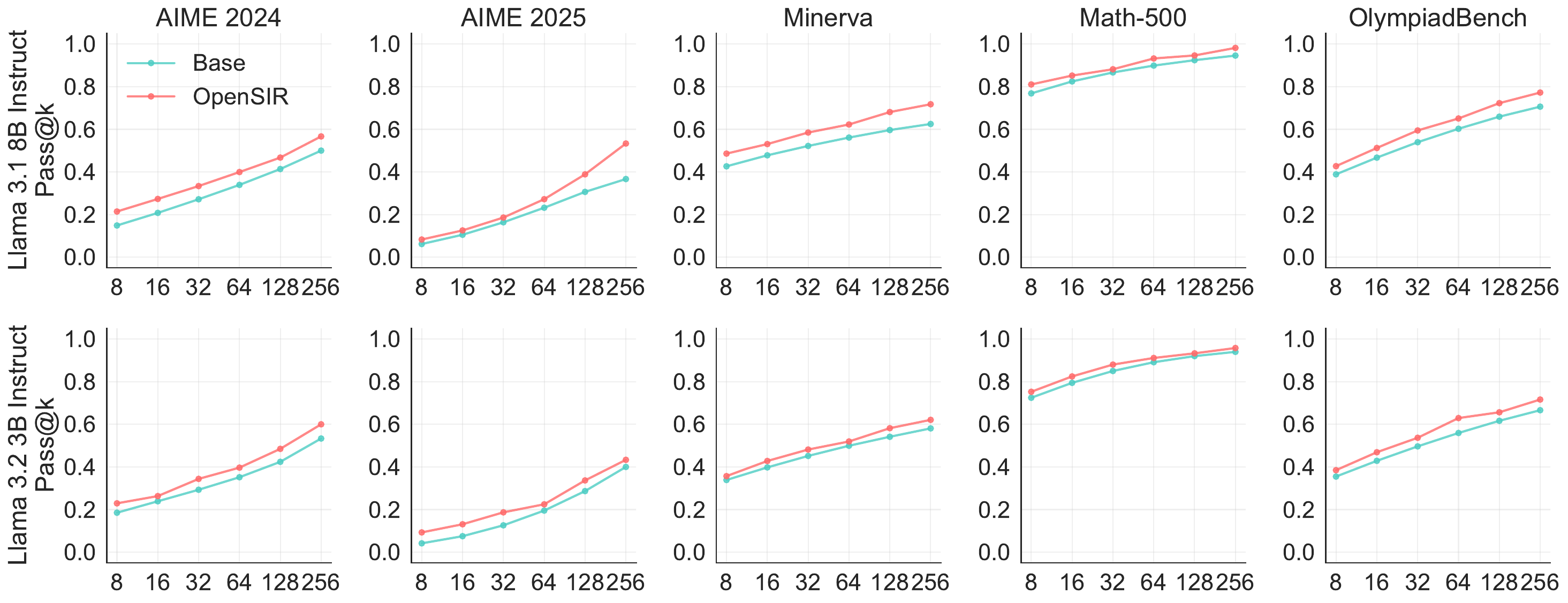}
  \caption{Pass@k curves comparing base instruction models and {\method} across five mathematical benchmarks. {\method} consistently improves performance across all k values, with stable or increasing gaps at higher k, demonstrating genuine reasoning improvements rather than memorization.}
  \label{fig:passk_curves}
\end{figure}

Figure~\ref{fig:passk_curves} shows that {\method} consistently outperforms base instruction models across all k values (8--256) on all benchmarks, with stable or increasing gaps at higher k.
These results suggest that {\method} improves the model's mathematical reasoning capacity.

\subsection{Prolonged Training Analysis}
\begin{figure}[htbp]

  \centering
  \includegraphics[width=.7\textwidth]{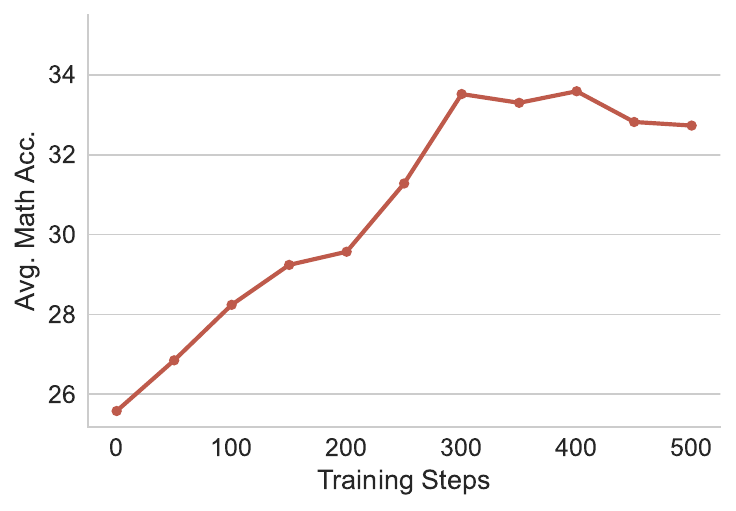}
  \caption{Performance of {\method} extended training using Llama-3.2-3B-Instruct.}
  \label{fig:opensir_extended}
\end{figure}
To better understand the limitation of {\method}, we extend training of a single run of Llama-3.2-3B-Instruct to 500 steps, substantially beyond the 200 steps used in the main experiments.
Figure \ref{fig:opensir_extended} shows the evaluation performances over training. We observe consistent improvement from 25.6\% to 33.5\% at step 300, representing a gain of +7.9 points. Performance then plateaus after step 300 and remains around 33\% through step 500, with no further noticeable gains in this single run.

By step 300, generated problems show narrower topical and difficulty variation than earlier checkpoints, and performance plateaus thereafter. Future work could explore richer seed curricula, larger model capacity, or complementary exploration strategies to prolong broad exploration before plateauing.

\subsection{Synergy with Annotated Data}

\begin{table}[!th]
  \centering
  \small
  \setlength{\tabcolsep}{3pt}
  \footnotesize
  \setlength{\tabcolsep}{3pt}
  \begin{tabular}{l|ccccccc|c}
    \toprule
    \multirow{2}{*}{\textbf{Model}}                        & \multirow{2}{*}{\textbf{GSM8K}} & \textbf{MATH}  & \multirow{2}{*}{\textbf{Minerva}} & \textbf{College} & \multirow{2}{*}{\textbf{Olymp.}} & \textbf{AIME} & \textbf{AIME} & \textbf{Math}        \\
                                                           &                                 & \textbf{500}   &                                   & \textbf{Math}    &                                  & \textbf{24}   & \textbf{25}   & \textbf{Avg.}        \\
    \midrule
    Base                                                   & 73.94                           & 42.86          & 15.21                             & 28.78            & 13.09                            & 5.00          & 0.21          & 25.58                \\
    GRPO$_{\text{gsm8k}}$                                  & 79.72                           & 45.30          & 16.27                             & 33.33            & 14.56                            & 6.82          & 1.88          & \cimp{28.27}{+2.69}  \\
    \rowcolor{lightblue!50}  {\method}                     & 78.28                           & 46.22          & 18.46                             & 35.42            & 16.72                            & 7.94          & 3.97          & \cimp{29.57}{+3.99}  \\
    \rowcolor{lightblue!50}  {GSM8K $\rightarrow$ \method} & 81.43                           & 46.12          & 19.43                             & 36.15            & \textbf{18.35}                   & 8.38          & 4.12          & \cimp{30.57}{+4.99}  \\
    \rowcolor{lightblue!50}  {GSM8K \& \method}            & \textbf{81.57}                  & \textbf{49.48} & \textbf{20.39}                    & \textbf{36.85}   & 18.14                            & \textbf{8.64} & \textbf{4.38} & \bcimp{31.35}{+5.77} \\
    \bottomrule
  \end{tabular}
  \caption{The Pass@1 performance on seven mathematical benchmarks. {\method} obtains better results when trained together with GSM8K compared to {\method} or GSM8K alone.}
  \label{tab:synergy-results}
\end{table}

While we showed that {\method} achieves significant improvements in math reasoning without using annotated data, we further investigate if {\method} can be combined with annotated data to achieve even greater performance gains.
Having demonstrated that {\method} achieves significant improvements without annotated data, we investigate whether combining {\method} with annotated data can yield further gains. We focus on Llama-3.2-3B-Instruct and use Gsm8K as the training data, as Table \ref{tab:math-results} shows that fine-tuning on GSM8K consistently outperforms using MATH.

We explore two training strategies: (1) \textbf{GSM8K $\rightarrow$ {\method}}: The model is first trained on GSM8K for half the training iterations, then trained with {\method} for the remaining half.
(2) \textbf{GSM8K \& {\method}}: Each training iteration uses half GSM8k samples and half {\method} samples.

Table \ref{tab:synergy-results} shows that both setups achieve better performance than using {\method} alone or only on GSM8K.
One possible explanation for the sequential approach's effectiveness (GSM8K $\rightarrow$ {\method}) is that training on GSM8K first may improve the model's foundational reasoning abilities, which could provide a stronger starting point for {\method}'s self-generated questions.
The concurrent approach (GSM8K \& {\method}) achieves a slight additional edge, which might be attributed to the model receiving better feedback signals for question calibration from the beginning, as it can leverage both supervised and self-generated data simultaneously throughout training.
The precise underlying mechanisms for these improvements require further investigation.

\subsection{Computational Cost Analysis}

Let $B$ be the rollout batch size per gradient update. {\method} uses double the rollout batch size of standard GRPO, split evenly between teacher (problem-generation) and student (solution) rollouts after problem selection (Section~\ref{sec:problem-selection}). The per-update cost is therefore exactly $2\times$ the baseline.

Problem embeddings for diversity scoring are computed asynchronously during solution generation, adding negligible wall-clock time. Cosine distance calculations require $\mathcal{O}(B \times |\mathcal{P}_t|)$ operations, where $|\mathcal{P}_t|$ is the problem pool size, and take under 3 seconds per iteration.

\section{Algorithm}
\label{sec:algorithm}

\begin{algorithm}[H]
  \caption{{\method}}
  \label{alg:opensir}
  \small
  \begin{algorithmic}
    \REQUIRE Problem pool $\mathcal{P}_0$, policy $\pi_\theta^{(0)}$, embedding encoder $\varepsilon$, batch size $B$, generation group size $G$, solve rate range $[s_{\min}, s_{\max}]$, teacher prompt $p_T$, student prompt $p_S$
    \FOR{$t = 1$ {\bfseries to} $T$}
    \STATE \textit{// Problem Generation}
    \STATE Sample $k = B/G$ reference problems $\{p_1, \ldots, p_k\}$ from $\mathcal{P}_{t-1}$
    \FOR{$i = 1$ {\bfseries to} $k$}
    \STATE Sample $q_{i,1:G} \sim \pi_\theta^{(t)}(\cdot \mid p_i, p_T)$
    \ENDFOR
    \STATE $\mathcal{Q}_{\text{valid}} \leftarrow \{q_{i,j} \mid q_{i,j} \text{ has valid format}\}$
    \STATE \textit{// Solution Sampling}
    \FOR{each $q_i \in \mathcal{Q}_{\text{valid}}$}
    \STATE Sample solutions $o_{i,1:G} \sim \pi_\theta^{(t)}(\cdot \mid q_i, p_S)$
    \STATE Parse answers $a_{i,1:G}$ from solutions $o_{i,1:G}$
    \STATE $a_i^* \leftarrow \arg\max_{a \in a_{i,1:G}} \text{count}(a)$ \textit{// reference answer via majority voting}
    \STATE Compute solve rate $s_{q_i} = \text{count}(a_i^*) / G$
    \STATE Compute embedding $e_{q_i} \leftarrow \varepsilon(q_i)$
    \ENDFOR
    \STATE \textit{// Scoring}
    \STATE Compute $\text{score}_{\text{novel}}(q_i)$ for all $q_i \in \mathcal{Q}_{\text{valid}}$ via Eq.~\ref{eq:novelty}
    \STATE \textit{// Teacher sample selection}
    \STATE $\mathcal{I}_T \leftarrow \text{top}_{B/(2G)}(i : \text{Var}(\text{score}_{\text{novel}}(q_{i,1:G})), i \in \{1,\ldots,k\})$
    \STATE \textit{// Student sample selection}
    \STATE $\mathcal{Q}_S \leftarrow \text{top}_{B/(2G)}(q : \text{score}_{\text{novel}}(q), q \in \mathcal{Q}_{\text{valid}})$
    \STATE Compute $\text{score}_{\text{correct}}(o_{i,j}, a_{i,j})$ for solutions where $q_i \in \mathcal{Q}_S$ via Eq.~\ref{eq:correct}
    \STATE \textit{// Model Update}
    \STATE $\mathcal{D}_T \leftarrow \{(p_T, q_{i,j}, R_{i,j}^T) : i \in \mathcal{I}_T, 1 \leq j \leq G\}$ where $R_{i,j}^T = \text{score}_{\text{novel}}(q_{i,j})$
    \STATE $\mathcal{D}_S \leftarrow \{(p_S, o_{i,j}, R_{i,j}^S) : q_i \in \mathcal{Q}_S, 1 \leq j \leq G\}$ where $R_{i,j}^S = \text{score}_{\text{correct}}(o_{i,j}, a_{i,j})$
    \STATE Update $\pi_\theta^{(t+1)} \leftarrow \text{GRPO}(\pi_\theta^{(t)}, \mathcal{D}_T \cup \mathcal{D}_S)$
    \STATE $\mathcal{P}_t \leftarrow \mathcal{P}_{t-1} \cup \mathcal{Q}_{\text{valid}}$
    \ENDFOR
    \STATE \textbf{return} $\pi_\theta^{(T)}$
  \end{algorithmic}
\end{algorithm}

\section{Annotation Details}
One of the authors prepared the samples for annotation, and the rest of the authors annotated the samples with the instructions provide in Figure \ref{fig:annotate_prompt}.
\label{sec:annotation_details}
\begin{tcolorbox}[breakable,colback=blue!5!white,colframe=blue!75!black,fontupper=\footnotesize,fontlower=\small,width=\textwidth ]
  You will be presented with multiple sets of 5 math problems to evaluate. For each set, please complete the following three-step annotation process.

  \# Step 1: Identify Topics

  For \textbf{each problem}, identify ALL relevant mathematical topics from the following list:

  - Algebra

  - Geometry

  - Calculus

  - Probability

  - Statistics

  - Number Theory

  - Combinatorics

  - Optimization

  - Arithmetic

  - Discrete Math

  - Trigonometry

  \# Step 2: Assess Validity

  For \textbf{each problem}, determine if it is \textbf{valid} or \textbf{invalid}:

  - \textbf{Valid}: The problem is logically sound, clearly stated, and can be answered with the given information

  - \textbf{Invalid}: The problem contains logical flaws, contradictions, insufficient information, or ambiguities that prevent a proper solution

  \# Step 3: Rank Difficulty

  Rank all 5 problems from \textbf{easiest to hardest}. Provide your ranking as a sequence of problem numbers.

  \textit{Example:} [3, 1, 5, 2, 4] means problem 3 is the easiest and 4 is the hardest.

  \textit{Consider these factors when assessing difficulty:}

  - Number of steps required

  - Complexity of concepts involved

  - Level of mathematical knowledge needed

  - Computational complexity

  \# Response Format

  Provide your annotations as a JSON list where each element represents one problem set. Here are some examples:

  \begin{verbatim}
[
  {
    "set_id": "SET_1",
    "problems": {
      "1": {"topics": ["Algebra", "Calculus"], "valid": true},
      "2": {"topics": ["Geometry"], "valid": false},
      "3": {"topics": ["Probability"], "valid": true},
      "4": {"topics": ["Number Theory"], "valid": true},
      "5": {"topics": ["Arithmetic"], "valid": true}
    },
    "difficulty_ranking": [5, 3, 1, 2, 4]
  },
  {
    "set_id": "SET_2",
    "problems": {
      "1": {"topics": ["Statistics"], "valid": true},
      "2": {"topics": ["Discrete Math"], "valid": true},
      "3": {"topics": ["Optimization"], "valid": true},
      "4": {"topics": ["Algebra"], "valid": false},
      "5": {"topics": ["Geometry", "Algebra"], "valid": true}
    },
    "difficulty_ranking": [1, 2, 5, 3, 4]
  },
  ...
]
    \end{verbatim}

\end{tcolorbox}
\captionof{figure}{The instruction provided to the annotators to annotate problems.}
\label{fig:annotate_prompt}

\paragraph{Inter-annotator agreement.}
To assess reliability, annotators shared a common set of 20 overlapping samples. We report Fleiss' $\kappa$~\citep{fleiss1971measuring} for topic labels ($\kappa = 0.82$) and validity ($\kappa = 0.86$), indicating satisfactory agreement on both categorical tasks. For difficulty rankings, we compute Kendall's coefficient of concordance $W$~\citep{kendall1939problem} per set and average across sets, obtaining $W = 0.67$, which reflects moderate alignment on relative problem difficulty.

\FloatBarrier
\section{Additional Ablations}
\subsection{Solution Length Reward Increases Problem Complexity}
\label{sec:sol-len}
\begin{table}[h]
  \centering
  \small
  \begin{tabular}{l|c|c|c}
    \toprule
    \multirow{2}{*}{\textbf{Model}} & \textbf{Question} & \textbf{Solution} & \textbf{Math}  \\
                                    & \textbf{Length}   & \textbf{Length}   & \textbf{Avg.}  \\
    \midrule
    w/ length                       & 207               & 387               & \textbf{29.57} \\
    w/o length                      & 150               & 238               & \textbf{28.09} \\
    \bottomrule
  \end{tabular}
  \caption{Comparison of {\method} performance with and without solution length reward. Solution length reward improves {\method} accuracy and increases average question and solution lengths.}
  \label{tab:solution-length-diversity}
\end{table}

We investigate the impact of the solution length reward in {\method}. Table~\ref{tab:solution-length-diversity} shows this reward improves performance from 28.09\% to 29.57\%.
It also increases the average question length (from 150 to 207 tokens) and solution lengths (from 238 to 387 tokens).
By examining the generated questions manually, we find that the policy tends to generate more sophisticated problems involving advanced concepts with this reward, such as linear programming and optimization, which naturally require longer multi-step solutions to solve. These results demonstrate that the solution length reward effectively guides the policy toward generating more complex problems, which in turn leads to better performance.

\FloatBarrier
\section{Implementation Details}
\subsection{Training Details}
\label{sec:train_details}
\begin{table}[h]
    \centering
    \begin{threeparttable}
        \begin{tabular}{lll}
            \toprule
            \textbf{Category} & \textbf{Hyperparameter}            & \textbf{Value}                          \\
            \midrule
            \multirow{8}{*}{Trainer}
                              & Learning rate                      & $3 \times 10^{-7}$                      \\
                              & Optimiser                          & AdamW \citep{loshchilov_decoupled_2018} \\
                              & Warmup steps                       & 20                                      \\
                              & Training steps                     & 200                                     \\
                              & KL loss coefficient                & $1 \times 10^{-4}$                      \\
                              & Gradient norm clipping             & 0.5                                     \\
                              & Seeds                              & 42/43/44                                \\
                              & GPUs                               & 3 H100                                  \\
            \midrule
            \multirow{5}{*}{Rollout}
                              & Batch size\tnote{$\dagger$}        & 256 / 512\tnote{$\ddagger$}             \\
                              & Max prompt length                  & 1024                                    \\
                              & Max solution length                & 2048                                    \\
                              & Number of rollouts per prompt      & 8                                       \\
                              & Temperature                        & 1.0                                     \\
            \midrule
            \multirow{5}{*}{Teacher Rewards}
                              & Solvability weight ($\alpha$)      & 1.0                                     \\
                              & Solution length weight ($\lambda$) & 1.0                                     \\
                              & Diversity  weight ($\gamma$)       & 1.0                                     \\
                              & Format  weight  ($\delta$)         & 0.1                                     \\
                              & Embedding model                    & Linq-Embed-Mistral (7B)                 \\
            \midrule
            \multirow{2}{*}{Student Rewards}
                              & Accuracy  weight                   & 1.0                                     \\
                              & Format  weight   ($\delta$)        & 0.1                                     \\
            \bottomrule
        \end{tabular}
        \begin{tablenotes}
            \footnotesize
            \item[$\dagger$] The number of rollouts seen for one gradient update.
            \item[$\ddagger$] Baselines use batch size 256; {\method} uses 512 because each batch is split between teacher (problem generator) and student (solver) rollouts, keeping the student-side problem-solution pairs per gradient step equal across methods.
        \end{tablenotes}
        \caption{The training configurations for the experiments.}
        \label{tab:train-details}
    \end{threeparttable}
\end{table}
We implement {\method} based on the TRL framework \citep{vonwerra2022trl}.
Each {\method} batch is split between teacher (problem generator) and student (solver) rollouts, so we double {\method}'s total rollout batch size relative to the baselines; this keeps the number of student problem-solution pairs per gradient step (and the total over training) equal across methods.
Table \ref{tab:train-details} provides a summary of the training hyperparameters used in our experiments.

\subsection{Evaluation Details}
\label{sec:eval_details}
We use sampling temperature $0.6$ and top-p $0.95$ with maximum response length 4,096 tokens (38,912 for reasoning models, following \citet{qwen3_2025}). We report Pass@1, averaged over 16 independently sampled generations per problem. Answer extraction and comparison use the \texttt{math\_verify} library.

\subsection{Prompts}
We detailed the prompt for generating problems in Figure \ref{fig:gen_q_prompt} and solving problems in Figure \ref{fig:gen_sol_prompt}.

\begin{figure}[htbp]
  \centering
  \begin{tcolorbox}[colback=blue!5!white,colframe=blue!75!black,fontupper=\normalsize,fontlower=\small,width=\textwidth ]
    You are given a math problem: \{Problem\}

    Your task is to create a math problem that is conceptually different from the provided problem. The new problem must be answerable with a numerical value or mathematical expression.

    First, explain how your new problem differs conceptually from the original problem inside the <think>...</think> tags. Then, present your new problem inside the <problem>...</problem> tags. Finally, identify at most three math concepts required to solve your problem. Provide these concepts in a comma separated list inside the <concepts>...</concepts> tags.

  \end{tcolorbox}
  \caption{Prompt for generating math problems. \{Problem\} is a placeholder for the reference problem sampled from the problem pool.}
  \label{fig:gen_q_prompt}
\end{figure}

\begin{figure}[htbp]
  \centering
  \begin{tcolorbox}[colback=blue!5!white,colframe=blue!75!black,fontupper=\normalsize,fontlower=\small,width=\textwidth ]
    You are a helpful AI Assistant, designed to provide well-reasoned and detailed responses. You FIRST think about the reasoning process step by step and then provide the user with the answer. The last line of your response should be 'Therefore, the final answer is: \verb|$\boxed{ANSWER}$|' (without quotes) where ANSWER is just the final number or expression that solves the problem.

    \{Problem\}

  \end{tcolorbox}
  \caption{Prompt for generating solutions to math problems. \{Problem\} is a placeholder for the actual problem.}
  \label{fig:gen_sol_prompt}
\end{figure}



\end{document}